\newif\ifOneColumn
\ifOneColumn
  \documentclass[journal,12pt,onecolumn,draftclsnofoot]{IEEEtran}
\else
  \documentclass[journal,twocolumn]{IEEEtran}
\fi
\usepackage{amsmath,amsfonts}
\usepackage{algorithmic}
\usepackage{algorithm}
\usepackage{array}
\usepackage[caption=false,font=normalsize,labelfont=sf,textfont=sf]{subfig}
\usepackage{textcomp}
\usepackage{stfloats}
\usepackage{url}
\usepackage{verbatim}
\usepackage[pdftex]{graphicx}
\usepackage{cite}
\usepackage{bigstrut, multirow,xcolor}
\usepackage{epstopdf}
\begin{document}

%\title{Uncertainty Injection: A New Deep Learning Training Scheme for Robust Optimization}
\title{Uncertainty Injection: A Deep Learning Method for Robust Optimization}

\author{\IEEEauthorblockN{Wei Cui, \IEEEmembership{Graduate Student Member,~IEEE}, and Wei Yu, \IEEEmembership{Fellow,~IEEE}}  % <-this % stops a space
\thanks{Manuscript submitted on March 5th, 2022. The materials in this paper have been presented in part at the IEEE International Conference on Acoustics, Speech, and Signal Processing (ICASSP),
Barcelona, Spain, May 2020 \cite{icassp}. This work was supported by Natural Sciences and Engineering Research Council (NSERC) of Canada.  
The authors are with The
Edward S.~Rogers Sr.~Department of Electrical and Computer Engineering,
University of Toronto, Toronto, ON M5S 3G4, Canada 
(e-mails: \{cuiwei2, weiyu\}@ece.utoronto.ca).}
}

% [New] commented this block

% [New] commented this block

\maketitle

\begin{abstract}
This paper proposes a paradigm of uncertainty injection for training deep
learning model to solve robust optimization problems. The majority of
existing studies on deep learning focus on the model learning capability, while
assuming the quality and accuracy of the inputs data can be guaranteed.
However, in realistic applications of deep learning for solving optimization
problems, the accuracy of inputs, which are the problem parameters in this
case, plays a large role. This is because, in many situations, it is often costly or sometime impossible to obtain the problem parameters accurately, and correspondingly, it is highly desirable to develop learning algorithms that can account for the
uncertainties in the input and produce solutions that are
\emph{robust} against these uncertainties. This paper presents a novel
\emph{uncertainty injection} scheme for training machine learning models that
are capable of implicitly accounting for the uncertainties and producing
statistically robust solutions. We further identify the wireless communications
as an application field where uncertainties are prevalent in problem parameters 
such as the channel coefficients. We show the effectiveness of the proposed training
scheme in two applications: the robust power loading for multiuser
multiple-input-multiple-output (MIMO) downlink transmissions; and the
robust power control for device-to-device (D2D) networks.  

\end{abstract}


%%%%%%%%%%%%%%%%%%%%%%%%%%%%%%
%%%%%Introduction Start%%%%%%%
%%%%%%%%%%%%%%%%%%%%%%%%%%%%%%
\section{Introduction}\label{sec:intro}

\IEEEPARstart{D}{eep} learning has achieved excellent results in a variety 
of different optimization tasks, such as the inference problems
\cite{alexnet,maas,resnet} and the generative modeling problems
\cite{vae,gan,radford}. Although the traditional domain of 
deep learning has been for applications in which the optimization 
problems do not admit explicit mathematical models, several recent advances
have also shown the promise of deep learning in solving non-convex 
optimization problems for which explicit mathematical formulations are 
available, e.g., \cite{unfold, learnwmmse, jsac}. 
This paper focuses on this latter class of problems 
in which a deep learning model is trained to take the 
parameters of the mathematical optimization problem as the input and to output
the optimized solution of the problem. Specifically, we focus on how to train
the model to achieve \emph{robustness} against uncertainty in the problem
parameters.

While most of the deep learning literature focuses on the learning performance 
of the neural network on a given dataset where the availability of 
high-quality input data is assumed, in many realistic 
applications, the uncertainties are inevitably part of the input (or the 
label data). When uncertainties are present, the performance of a solution
under the uncertain input realizations is often of importance, and this is 
commonly referred to as the robustness of the solution \cite{szegedy, fawzi}. %While a variety of literature already exists exploring the robustness of deep learning algorithms, research along this topic is still relatively under explored, leaving much room for novelties. 

Specifically, several research directions on the robustness of deep learning 
have been investigated in the literature. 
In \cite{natarajan,joulin,veit,rolnick}, the performances of supervised 
learning when trained with uncertainty in the targets are investigated. 
Meanwhile, the robustness of deep learning models on 
\emph{data distribution changes} has also been explored, often referred to as 
\emph{distributional robustness}, as in \cite{sinha,pleiss,duchi,hein}. 
In \cite{jianping,ordonez,yanxia,shixiang}, the input cleaning procedure is
utilized for deep learning models given initially noisy inputs.
Besides the above-mentioned approaches, researchers have also explored training
deep learning models with noise actively added at the inputs
\cite{bishop,neuralsmithing,shaham,moosavi,madry}, into the neural network
parameters \cite{graves}, into the activation functions \cite{benpoole,caglar},
and into the gradients \cite{neelakantan,mozhou}. These classes of techniques 
have 
been shown to improve the robustness of the deep learning models against small
perturbations or adversarial attacks in testing inputs, and to encourage the
models to produce better generalization results. 

Despite the large number of robust deep learning research works as mentioned
above, most of them deal with problems for which explicit mathematical models 
do not exist, and none specifically target towards solving mathematical 
non-convex robust optimization problems. For example, although elaborated
non-convex optimization techniques and analysis have been used for obtaining
robust deep learning models in works such as \cite{fawzi,shaham}, they are 
not designed to obtain deep learning models specifically to take the uncertain 
problem parameters as the input and to produce robust solutions to a
mathematical non-convex optimization problem as the output.

Robust optimization has been extensively studied in the traditional
mathematical programming literature. 
%Meanwhile, robustness in mathematical non-convex optimization is particularly prevalent in many fields. One such 
For example, in the area of wireless communication network utility 
maximization, the optimization of network operations typically involves
first obtaining wireless network parameters such as the channel state
information (CSI), then formulating a network utility objective as a function
of these network parameters, and finally optimizing the objective function
assuming these fixed parameters \cite{FlashLinQ, shen_ISIT17, luo_TSP11, ITLinQ,
Guo_TCOM17, color, MAPEL, Johansson_TWC06}. This {deterministic
optimization} framework may not always produce the best solution in
realistic situations, because it inherently ignores the channel uncertainties,
which can significantly affect the quality of the solutions. On the other
hand, researchers have explored mathematical robust optimization techniques
that incorporate these uncertainties. The classical approaches for dealing
with wireless channel uncertainty within the optimization process either 
assume bounded uncertainty regions \cite{shenouda,monowar,shenkwak,weihua}, 
or incorporate statistical models of channel uncertainty
\cite{junwang,dallanese,chalise,shenouda2,kunyu,foad,medra,elnourani}. Although fitting reality better than
deterministic optimization, these robust optimization approaches rely
on the mathematical models of the uncertainty in the parameters, which are
often ad-hoc. Further, the parameters of these models are not easy to
estimate. Finally, even if the model and its parameters are known exactly, the
resulting optimization problem is often difficult to solve.  We note that there
has been several work on using deep learning for obtaining robust solutions to
wireless communication problems \cite{kerret,junbeom,runze}, however these
work only focus on the expectation of the achievable rates under channel
uncertainty and do not fully capture the notion of robustness from statistical
distribution point of view. % (e.g., robust rate under a 5-percentile outage constraint).

% under the context of the optimization problems in these works are either not well formulated, or are too simple (i.e. only focusing on the expectation under uncertainties).

This paper proposes a novel deep neural network training strategy for
maximizing a statistical robustness measure for 
non-convex utility optimization problems, addressing gaps in both
deep learning and wireless communication research. We advocate statistical
uncertainty models rather than bounded-region uncertainty models due to the
fact that realistic uncertainties are generally not guaranteed to be bounded.
But instead of relying on the mathematical representations of the
statistical distribution of the uncertainties, we pursue a data-driven 
approach to robust optimization, because it is usually much more practical
to obtain samples of the uncertainty realizations rather their mathematical 
representations. 

The key innovation of this paper is that we take full advantage of the fact
that we are solving an explicitly formulated mathematical programming problem
by feeding an estimate of the problem parameters as input, then obtaining the
optimized solution as the output of the neural network. In this case, the
optimized solution at the output can be further evaluated under 
the parameter uncertainties.
This allows us to propose a novel training strategy for deep learning of
directly injecting the parameter uncertainty samples \emph{after the output
layer of the neural network} to obtain the robust objective, then optimizing
the neural network weights using the gradients computed from the robust
objective. Because neural networks are universal and highly flexible function approximators, a neural network trained under these parameter uncertainty
samples can implicitly infer the uncertainty distribution, thus
producing optimized solutions that are robust against the uncertainties in the
problem parameters.

To illustrate the effectiveness of the proposed training scheme, we focus on
two wireless network optimization problems under the robust {minimum-rate}
maximization objective: power loading for the multiuser
multiple-input-multiple-output (MIMO) downlink channel, and power control for
wireless device-to-device (D2D) networks. The sources of parameter uncertainties
are channel estimation error and the fading in wireless channels. 
Under the statistical uncertainty model,
similar to that of \cite{junwang,dallanese}, we adopt an \emph{outage-based}
notion of robustness in the optimization formulation. The minimum-rate
is adopted as the objective for both problems due to its emphasis on the
\emph{fairness} among the transmission links. We show in this paper that
uncertainty injection at the output can significantly improve the robustness of
the power allocation.

To summarize, the main contributions of the paper are as follows:
\begin{itemize}
\item We recognize the difficulties in the state-of-the-art robust optimization studies in terms of the uncertainty modeling and algorithmic complexity. 
\item We advocate a sample-based parameter uncertainty characterization for greater generalization ability, higher expressive power, and simplicity. 
\item We propose a novel deep learning based robust optimization scheme by uncertainty injection at the neural network's output layer, through which the neural network can be trained to perform robust optimization using only samples of the parameter uncertainties. 
\item We illustrate the effectiveness of the proposed approach for two important wireless communication applications: robust power loading in MIMO downlink networks, and robust power control in D2D networks.
\end{itemize}

The rest of this paper is organized as follows. Sections~\ref{sec:prob} and \ref{sec:prob_wireless}
formulate the general robust optimization framework with deep learning, along
with two wireless network optimization problem settings.
Section~\ref{sec:method} proposes a novel sample-based uncertainty-injection training scheme for deep learning to produce statistically robust solutions. 
The application of the proposed method to wireless communications is described
in Section~\ref{sec:app_wireless} and 
its performance is analyzed in Section~\ref{sec:exp}. Finally, conclusions are drawn in Section~\ref{sec:conclusion}.

%%%%%%%%%%%%%%%%%%%%%%%%%%%%%%%%%%
%%%%%Problem Formulation Start%%%%
%%%%%%%%%%%%%%%%%%%%%%%%%%%%%%%%%%
\section{Robust Optimization Formulation}\label{sec:prob}

We first present the general mathematical formulation of non-convex utility
optimization problem, in which the notion of robustness is defined under the
statistical distribution of the parameter uncertainties.

\subsection{Problem Setup}\label{sec:prob_I}
Consider a general optimization problem $\mathcal{P}$, consisting of the following components:
\begin{itemize}
    \item True problem parameters $\mathbf{p}$ summarizing all the information about the environment (but not perfectly known);
    \item Measurements $\mathbf{q}$ about the problem parameters;
    \item %Provided only with $\mathbf{q}$, 
	Optimization variables $\mathbf{x}$;
    \item Scalar utility function $u_{\mathbf{p}}(\mathbf{x})$ of 
	the optimization problem, as computed
	at the optimization variables $\mathbf{x}$, 
	assuming that the problem parameters are $\mathbf{p}$. 
	Here, $u_{\mathbf{p}}(\mathbf{x})$ is assumed to be differentiable 
	in $\mathbf{x}$ almost everywhere.
\end{itemize}
% (where the superscript denotes the utility value achieved under the set of parameters $\mathbf{P}$).
%We assume the objective function $u(\cdot)$ to be a scalar function and is differentiable over its argument $\mathbf{x}$. %Both assumptions are easily satisfied and are applicable to a wide range of realistic optimization applications.
The goal of robust optimization is to find an optimized $\mathbf{x}$, 
based on $\mathbf{q}$, that maximizes a robust objective of the utility function 
under some joint statistical 
distribution of the problem parameters $\mathbf{p}$ and the measurements $\mathbf{q}$.

\subsection{Statistical Distribution of Problem Parameters}\label{sec:prob_II}

In the existing robust optimization literature, there are two main parameter 
uncertainty models:
%When exploring robustness of optimization solutions against parameter uncertainties, two main parameter uncertainty models exist in the literature: 
\begin{enumerate}
    \item The uncertain parameters are confined within some closed sets, with ellipsoids being the most popular choice. The corresponding notion of robust objective is typically the worst-case utility value over the parameters in the uncertainty sets. \label{robustnotion_1}
    \item Parameter uncertainties follow some statistical distribution, with well-studied distributions (e.g. exponential family) being the most popular choices due to feasibility of the ensuing mathematical optimization. The corresponding robust objective is typically a statistical measure of the resultant utility, e.g., the 5th-percentile outage value. \label{robustnotion_2} 
\end{enumerate}
%While \ref{robustnotion_1}) is relatively more popular in existing literature, the assumption of uncertainties being bounded is unfitting for many natural phenomenons (a trivial example would be Gaussian distributed noises or errors).
%The notion of robustness associated with bounded uncertainties is the worst-case performance of the solution. Such robust solutions might not inefficient or over-conserved, since the corresponding worst-case scenario might occur with extremely low probabilities.
The bounded uncertainty model is mathematically more tractable, but less well
justified in practice. Its associated worst-case performance also tends to be 
over-conservative, since the worst-case scenario may occur only with very low
probability. For this reason, this paper adopts the statistical model
for the parameter uncertainties. The statistical uncertainty model lends 
well to data-driven approaches in which samples of the uncertain parameters 
can be generated and the corresponding optimization objective can be estimated
empirically under these samples.

%With the above-mentioned reasons, we adopt \ref{robustnotion_2}) as our uncertainty model. Furthermore, we relax the constraint on the statistical distributions being well-studied as required by existing traditional mathematical optimization literature. As we will show in the following sections, our method for robust optimization only relies on samples from the uncertainty distributions, and therefore reducing the need for the distributions themselves to be with manageable complexities.

We note that there are existing works in the literature such as 
\cite{monowar,weihua} that assume a
bounded uncertainty model, while defining the robust objective
probabilistically.  These works rely on bounding the statistical quantities
using often complex and mathematically involved derivations and approximations.
The point of this paper is that instead of attempting to approximate these
statistical quantities analytically, we use a data-driven approach to evaluate
the desired robust objective empirically.

%They made the bridging between the uncertainty model to the notion of robustness via finding proper values of bounds for the modeled uncertainties. These bounds would correspond to certain probabilities of uncertainty realizations, and consequentially lead to probabilities on the objective values. However, to obtain such bounds, complex mathematical derivations and heavy approximations are often needed on the uncertainties and the objective functions, which again defy our purpose of developing a robust optimization algorithm for general and potentially mathematically intractable uncertainties.

%We note that there exists works such as \cite{monowar,weihua} that assume bounded uncertainty model \ref{robustnotion_1}), while defining the robustness of the solution by probabilities. They made the bridging between the uncertainty model to the notion of robustness via finding proper values of bounds for the modeled uncertainties. These bounds would correspond to certain probabilities of uncertainty realizations, and consequentially lead to probabilities on the objective values. However, to obtain such bounds, complex mathematical derivations and heavy approximations are often needed on the uncertainties and the objective functions, which again defy our purpose of developing a robust optimization algorithm for general and potentially mathematically intractable uncertainties.

%Therefore, we adopt statistical distributions to model immeasurable uncertainties within the inputs. We formalize such 

Toward this end, we define the following statistical model on the problem
parameter $\mathbf{p}$ given the measurements of the environment $\mathbf{q}$ 
as follows:
 %by defining the distribution of possible real realizations of $\mathbf{P}$ given the measured or estimated input values $\mathbf{q}$:
\begin{align}
    \mathbf{p} \sim f_{\mathbf{p}|\mathbf{q}}(\mathbf{p}|\mathbf{q})
\end{align}
We emphasize that in the proposed data-driven approach, the above uncertainty distributions need not be expressed in analytic closed form. As we shall see later, the proposed method can be applied as long as we can obtain offline samples from the above distribution.

\subsection{Robust Objective under Statistical Uncertainty Distribution}\label{sec:III}

Given the measurement $\mathbf{q}$ and the corresponding distribution of the
problem parameters $\mathbf{p} \sim f_{\mathbf{p}|\mathbf{q}}(\mathbf{p}|\mathbf{q})$, 
the value of the utility function $u_{\mathbf{p}}(\mathbf{x})$ for any given
$\mathbf{x}$ would also follow some distribution. 
We define the robust objective of the optimization problem by the
\emph{percentile value} of the resulting distribution of
$u_{\mathbf{p}}(\mathbf{x})$. Specifically, define the $\gamma$-th percentile
value of the utility as the largest $u^\gamma$ for which 
\begin{align}
    \mathrm{Pr}[u_{\mathbf{p}}(\mathbf{x})\leq u^\gamma|\mathbf{q} ] \le \gamma\%.
\end{align}
The robust optimization problem $\mathcal{P}$ can now be formulated as%
\begin{subequations}
\label{robust_prob}
\begin{align}
\underset{\mathbf{x}}{\text{maximize}}\quad& u^\gamma\\
\text{subject to}\quad& \mathrm{Pr}[u_{\mathbf{p}}(\mathbf{x})\leq u^\gamma| \mathbf{q} ] \leq \gamma\% 
\end{align}
\end{subequations}
where the probability is taken under $\mathbf{p} \sim f_{\mathbf{p}|\mathbf{q}}(\mathbf{p}|\mathbf{q})$. The value $\gamma$ can be interpreted as an outage probability.

%The optimization solution $\mathbf{x}$ enjoys a notion of robustness, as the resultant objective is higher than the threshold $u^\gamma$ which is maximized, with probability at least $1-\frac{\gamma}{100}$. 

\subsection{Data-Driven Approach to Robust Optimization}\label{sec:IV}

\begin{figure*}[!h]
    \centering
    \ifOneColumn
      \includegraphics[width=\textwidth]{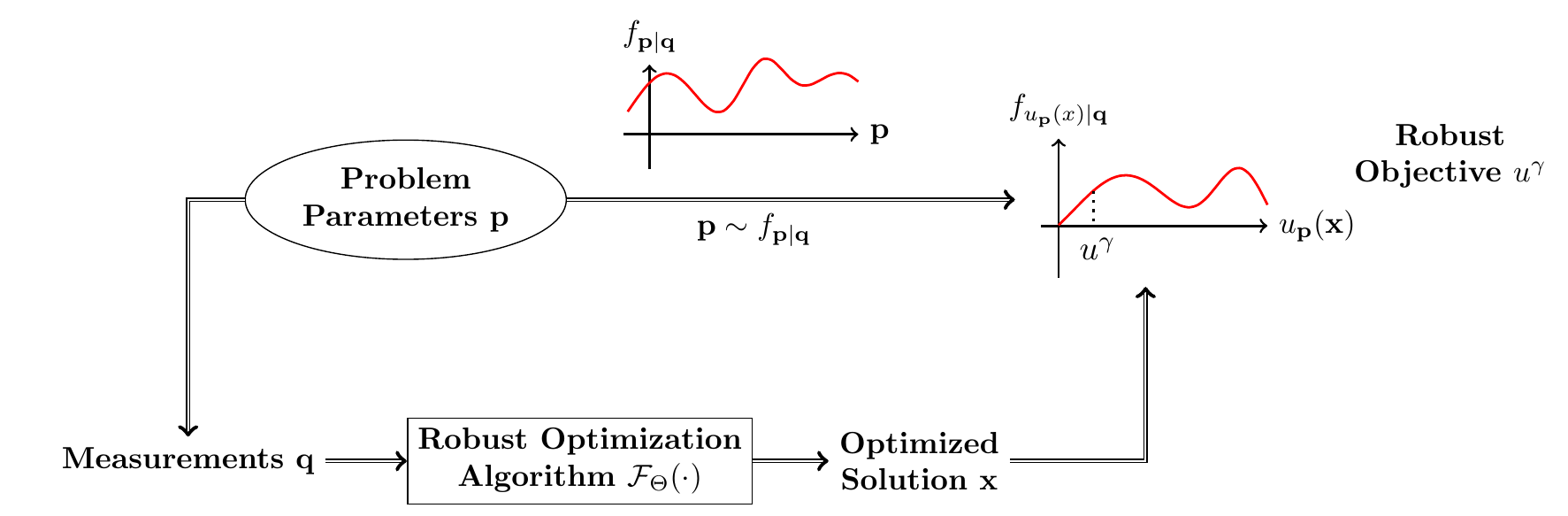}
    \else
      \includegraphics[width=0.8\textwidth]{fig/optimization_flow.pdf}
    \fi
    \caption{The robust optimization problem for maximizing a $\gamma$-percentile outage utility based on a measurement of the problem parameters.}
    \label{fig:optimization_flow}
\end{figure*}

This paper aims to solve the the robust optimization problem (\ref{robust_prob}) 
above based on the measurements $\mathbf{q}$ of the unknown problem parameters 
$\mathbf{p}$ by producing
an optimized variable $\mathbf{x}$ that maximizes the robust objective under 
the outage constraint. % using 
%optimizing the objective $u^{(\mathbf{P})}(\mathbf{x})$ while being robust against unknown uncertainties in the real parameters $\mathbf{P}$. 
%an optimization process as shown in Fig.~\ref{fig:optimization_flow}. 

In this end, this paper proposes to utilize a deep learning approach to map
the measurement of the optimization problem parameters to a robust optimized
solution. The deep neural network is chosen for its computation capacity and
representation ability. Using a neural network, the optimization algorithm can
be represented as:
\begin{align}\label{equ:neuralnet_mapping}
    \mathbf{x}={\mathcal F}_{\Theta}(\mathbf{q}) 
\end{align}
where $\Theta$ is the collection of the neural network model parameters and hyper-parameters. The optimization problem $\mathcal{P}$ then translates to finding a set of high-quality neural network parameters $\Theta$.
The overall optimization process is shown in Fig.~\ref{fig:optimization_flow}. 

\section{Robust Optimization in Wireless Communications and Networking}
\label{sec:prob_wireless}

We now present two applications of the above general robust optimization 
framework in wireless network utility maximization: the robust power 
loading problem for multiuser MIMO transmissions and the robust power control 
problem for D2D networks. Wireless network utility optimization problems naturally fit into the robust optimization framework, since the wireless channels (i.e. the problem parameters $\mathbf{p}$) are difficult to measure accurately, and the channel estimation process always produces measurement uncertainties.

\subsection{Robust Beamforming for Minimum Rate Maximization in Multiuser MIMO Downlink}\label{sec:prob_III}

\begin{figure}
    \centering
    \ifOneColumn
      \includegraphics[width=0.45\textwidth]{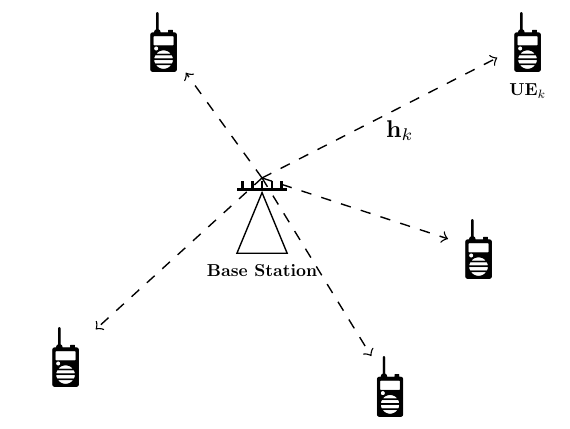}
    \else
      \includegraphics[width=0.3\textwidth]{fig/MIMO_Diagram.pdf}
    \fi
    \caption{Multiuser MIMO Downlink Channel}
    \label{fig:mimo_diagram}
\end{figure}

Consider a MIMO transmission scenario with one base station equipped with $M$ antennas
serving $K$ single-antenna users. The base station serves all $K$ users through
multiuser MIMO downlink transmission in the same time-frequency resource block. 
%s, with $p_k$ as the maximum power reserved for the $k$-th user. 
We use $\mathbf{H}=[\mathbf{h}_1,\dots,\mathbf{h}_k,\dots,\mathbf{h}_K]
\in\mathbb{C}^{M\times K}$ to denote the channel matrix, with its
$k$-th column $\mathbf{h}_k\in\mathbb{C}^{M\times1}$ denoting the channels from
the base station antennas to the $k$-th user. 
In practical wireless communication scenarios, $\mathbf{H}$ is often not known perfectly.
The goal is to design robust downlink beamforming vectors against the uncertainties in $\mathbf{H}$.
A diagram illustrating the downlink multiuser MIMO channel is shown in Fig.~\ref{fig:mimo_diagram}.

%We use $\sigma^2$ to denote the background noise power level. We assume full frequency reuse among all the uplink and downlink transmissions over bandwidth $w$. 

In a dense urban environment, there are often no line-of-sight (LoS) paths from
the base station to the users, and the wireless channels in the multiuser MIMO
networks can be modeled as a Rayleigh fading channel \cite{rayleigh}.  To
estimate the channel, pilot signals need to be used. For time-division duplex
(TDD) systems, uplink pilots from the users to the base station can be used to
estimate the uplink channel. Then, based on the uplink-downlink reciprocity,
the downlink channel can be inferred.  For frequency-division duplex (FDD)
systems, downlink pilots from the base station to the users need to be used. 
In this case, each user estimates its own channel, then feeds back a quantized
version of the channels to the base station in order to enable the base station
to design the downlink beamformers based on the estimated channels from all the users.

In either case, the accuracy of the estimated channels $\hat{\mathbf{H}}$ 
depends on the pilot length, the background noise level, and in the FDD case
is also a function of feedback rate. 
%Based on pilot signals, the base station can then obtain channel estimations using the minimum mean square error (MMSE) \cite{mmse_channel} estimation. 
A common way to model the channel estimation error is to assume that 
for the wireless channel from the $j$-th base station antenna to the $k$-th 
user, the relationship between the real
channel coefficient $h_{kj}$ and the estimated channel coefficient
$\hat{h}_{kj}$ can be modeled as the following: 
\begin{align}\label{equ:MIMO_uncertainty}
    h_{kj} = \hat{h}_{kj} + e_{kj},\; e_{kj}\sim\mathcal{CN}(0,\sigma_e^2) \quad \forall j,k
\end{align}
where the estimation error $e_{kj}$ is independent and identically distributed (i.i.d.) across all the channels.

The robust beamforming problem for the multiuser MIMO system is that of designing the beamformers at the base station, along with the power loading for each beamformer, in order to achieve a robust objective. In this paper, we treat the optimization objective of maximizing the minimum rate in order to provide fairness across all the users. Although ideally, one could conceivably design both the beamformers and the power loadings jointly to account for the channel uncertainty, such an approach would have been too challenging due to its analytic complexity. In the existing literature, only the robust optimization for the simpler objective of power minimization has been shown to be possible to solve analytically over both the beamformers and the power allocation variables \cite{chalise,shenouda2,kunyu}. In contrast, this paper adopts a different approach: we fix the beamformers and assume that the design of the beamformers can be done based on $\hat{\mathbf{H}}$ only, then rely on the subsequent power allocation to ensure robustness. This design approach can be an effective one due to that it significantly simplifies the overall design process, while still achieving competitive robust performances as shown in \cite{foad,medra}.

More specifically, based on the estimated channels $\hat{\mathbf{H}}$, the 
base station can apply any one of the well-established beamforming techniques, 
such as zero-forcing (ZF) or regularized zero-forcing (RZF) \cite{rci}, 
to design a fixed set of precoders $\mathbf{B}\in\mathbb{C}^{M\times K}$ for
downlink transmission. We use the notation
$\mathbf{B}=[\mathbf{b}_1,\dots,\mathbf{b}_k,\dots,\mathbf{b}_K]$, with the
$k$-th column $\mathbf{b}_k\in\mathbb{C}^{M\times1}$ with unit norm
$\|\mathbf{b}_k\|_2=1$ being the beamformer from the base station to transmit 
information to the $k$-th user. 

The robust optimization variables are now the set of power loading variables $\mathbf{x}=\{x_k\}_{k\in[1\dots K]}$, where $x_k \in[0, 1]$ denotes the proportion of total power the base station should allocate for transmitting to the $k$-th user for optimizing a network-wide utility $u_{\mathbf{p}}(x)$. In this paper, we use the minimum rate across all the users as the utility. Note that for given $\mathbf{x}$, the achievable rate for the $k$-th user is computed as
\begin{align} \label{equ:MIMORate}
r_k = w\log\left(1+\frac{P |\mathbf{h}_k^H\mathbf{b}_k|^2x_k }{P \sum_{j\neq k}|\mathbf{h}_k^H\mathbf{b}_j|^2 x_j + \sigma^2}\right),
\end{align}
where $P$ denotes the total power constraint, $w$ denotes the bandwidth, $\sigma^2$ denotes the background noise level, and $(\cdot)^H$ denotes Hermitian transpose. 

For any $\mathbf{x}$, the statistical channel uncertainties induce a distribution on the achievable rates of each user, and consequentially, a distribution on $r_{\min}$, the minimum rate among all the users. We can compute the percentile value $r_{\min}^\gamma$ of the distribution for $r_{\min}$ as the maximum value for which: 
\begin{equation}
	\mathrm{Pr}[r_{\min}<r_{\min}^\gamma|\hat{\mathbf{H}}] \leq \gamma\%.
\end{equation}

With these notations established, the multiuser MIMO robust beamforming problem is now readily formulated as an instance of the robust optimization problem $\mathcal{P}$ as in Section~\ref{sec:prob_II}, with the following correspondence:
% (where we omit the trivial correspondence between the power loading solutions and the optimization variables, as both are denoted by $\mathbf{x}$):
\begin{align}
    \mathbf{p} \leftarrow& \: \mathbf{H} \nonumber \\
    \mathbf{q} \leftarrow& \: \hat{\mathbf{H}} \nonumber \\
    u_{\mathbf{p}}({\mathbf x}) \leftarrow& \: r_{\min} \nonumber \\
    u^\gamma \leftarrow& \: r_{\min}^\gamma \nonumber
\end{align}

The \emph{robust power loading} problem for maximizing the minimum-rate in a multiuser MIMO downlink, which we denote by $\mathcal{P}_{\text{MIMO}}$, is then formulated as follows:
\begin{subequations}\label{robust_MIMO_prob}
\begin{align}
\underset{\mathbf{x}}{\text{maximize}}\quad&
r_{\min}^\gamma\\
\text{subject to}\quad& \mathrm{Pr}[r_{\min}<r_{\min}^\gamma|\hat{\mathbf H}]\leq\gamma\%\\
& r_{\min}\leq r_k,\;\forall k\\
& \sum_{k=1}^K x_k \leq 1, \quad x_k \geq 0,\;\forall k \label{equ:robust_MIMO_prob_I}\\
&\text{and }(\ref{equ:MIMO_uncertainty}), (\ref{equ:MIMORate}) .\nonumber
\end{align}
\end{subequations}

\subsection{Robust Power Control for Minimum Rate Maximization in D2D Wireless
Networks}\label{sec:prob_IV}

\begin{figure}
    \centering
    \ifOneColumn
      \includegraphics[width=0.5\textwidth]{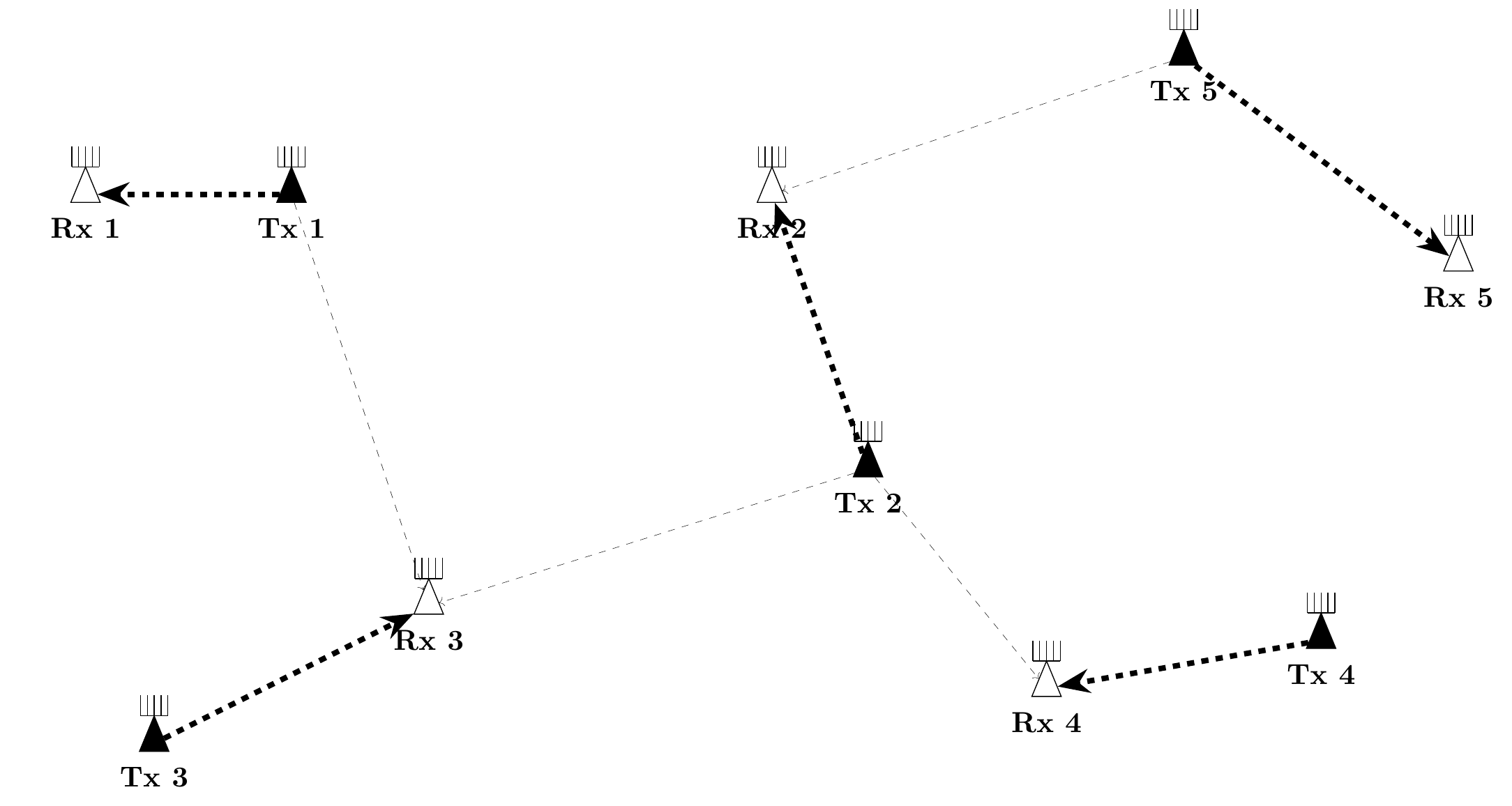}
    \else
      \includegraphics[width=0.4\textwidth]{fig/D2D_Diagram.pdf}
    \fi
    \caption{D2D Interfering Network}
    \label{fig:d2d_diagram}
\end{figure}

Consider a wireless ad-hoc network with $N$ independent D2D links with 
full frequency reuse in a two-dimensional region as illustrated in
Fig.~\ref{fig:d2d_diagram}. The goal is to find a set of power setting at 
each transmitter so as to be able to mutually accommodate the simultaneous
transmissions for all links. In this application, we again use the minimum
rate across all the users as the network utility function, in order to ensure
fairness.  This is a challenging task, because due to the aggressive frequency
reuse in the shared medium, the aggregate interference from the neighboring
links pose as significant impairments to each of the transmission pairs.

In the setting considered in this paper, we assume that the transmitters and
the receivers are equipped with multiple antennas, further the links operate 
in the millimeter wave (mmWave) frequency and there is a 
dominant LoS path between each transmitter and receiver pair. Moreover, we assume
that in the network deployment phase, a beam alignment procedure has taken
place between each transmitter-receiver pair, so that all the direct channels can
benefit from a substantial array gain. Then, an additional power
optimization step across all the users is performed 
to ensure that the aggregated interference is under control for
each transmission pair. Here, we focus on the power optimization step.

%Each transmitter and receiver is equipped with a linear antenna array with $M$ antennas. 

%We use $p_i$ for the maximum power at the $i$-th transmitter; $\mathbf{G}$ for the set of channel gains including all direct and interfering channels, with $g_{ij}\in\mathbb{R}$ being the channel gain from the $j$-th transmitter to the $i$-th receiver; and $\sigma^2$ for the background noise power level.  We assume full frequency reuse among all the links over bandwidth $w$. A diagram illustration for the D2D network, with direct links and a few selected cross links labelled, is shown at Figure~\ref{fig:d2d_diagram}.

To formulate and to solve a power optimization problem across $N$ transmitting
and receiving nodes using traditional mathematical programming technique, one
would need to estimate not only all the direct channels, but also all the
interfering channels between every transmitter and every receiver. In a network
of $N$ transmission pairs, one would need a dedicated pilot phase of duration at
least $O(N^2)$ in order to estimate all of the $N^2$ channel coefficients.
This is often infeasible. 

%Due to the highly aggressive frequency reuse in D2D wireless networks, if we utilize pilot signals for channel estimation as we did for the MIMO networks, we would need very long pilot sequence lengths to ensure the pilots are orthogonal to each other, which would incur high time and energy overhead. On the other hand, D2D networks consist of short links, where LOS paths are often available (especially when being deployed at rural areas). 

In this part of the paper, we explore the possibility of performing power 
control purely based on the geographic location information of all the
transmitters and the receivers. The geographic location information already 
provides the pathloss component of the overall channel. The idea is to formulate
a robust optimization problem, so that a reasonable minimum rate across all 
the users can still be achieved with high probability, even if the power control
is based only on the pathloss information.

%related path-loss components are dominant in the D2D wireless channels. As the result, instead of using pilots and obtain channel estimations as optimization inputs for the MIMO network scenario, we utilize the stable and readily obtainable GLI measurements to compute our channel estimates as inputs to the optimization problem. 
Another benefit of utilizing geographic location information is that such information can also be readily used for computing the beamformers at the transmitters and 
the receivers based on the angles of arrival and departure of the intended 
signal transmission to align the beams of the transmitter and the receiver 
towards each other in the initial deployment phase. 

In the ensuing channel model, we use $\hat{\mathbf G}^\mathrm{PL}$ to denote 
the path-losses of all the channels, including the beamforming gains at both
the transmitters and the receivers. In addition, we assume a log-normal
shadowing component and a fast-fading component that contribute towards the
uncertainties in the channel model. Specifically, we assume a log-normal
distribution for the shadowing $\mathbf{G}^\mathrm{S}$, and a circularly
symmetric complex Gaussian distribution for the fast fading coefficients that lead to
a fast fading component $\mathbf{G}^\mathrm{F}$. Then, the overall channel 
$\mathbf G$ has its $(i,j)$-th component distributed as: 
\begin{align}\label{equ:D2D_uncertainty}
    g_{ij} =&\: \hat{g}^\mathrm{PL}_{ij}g^\mathrm{S}_{ij} g^\mathrm{F}_{ij} 
\end{align}
with the log-normal shadowing component as
$$
g^\mathrm{S}_{ij} = 10^{\frac{g^{\mathrm S\text{dB}}_{ij}}{10}},  \qquad g^{\mathrm S\text{dB}}_{ij}\sim\mathcal{N}(0,\sigma_S^2), \qquad \forall i,j
$$
and the Rayleigh fading component as
$$
    g^\mathrm{F}_{ij}\sim\chi^2(2), \qquad \forall i,j 
$$
where $\sigma_S$ is the standard deviation of the shadowing in the dB scale, while the standard deviation of the circularly symmetric complex Gaussian distribution for the fast fading coefficients is assumed to be 1. Here, $\chi^2(2)$ denotes the chi-squared distribution with 2 degrees of freedom.

The robust optimization variables are the set of power control variables $\mathbf{x}=\{x_i\}_{i\in[1\dots N]}$, where $x_i \in[0, 1]$ denotes the proportion of total $p_i$ the $i$-th transmitter should transmit. Given $\mathbf{x}$, the achievable rate for the $i$-th link is computed as
\begin{align} \label{equ:D2DRate}
r_i = w\log\left(1+\frac{g_{ii}p_i x_i}{\sum_{j\neq i}g_{ij} p_jx_j + \sigma^2}\right),
\end{align}
where $w$ is the bandwidth, $p_i$ is the power budget of link $i$, and $\sigma^2$ is the background noise power.

Under a fixed power allocation $\mathbf{x}$, the statistical variations of the channel result in statistical variations in the achievable rate of each link. As a result, the minimum rate among all the links, i.e.,
\begin{equation}
	r_{\min} = \min_{i=1,\cdots,N} r_i
\end{equation}
also follows a distribution induced by the channel uncertainties. As in Section~\ref{sec:prob_II}, we adopt the $\gamma$-th percentile value of the minimum rate distribution, $r_{\min}^\gamma$, as the objective value, i.e.,  $r_{\min}^\gamma$ is the largest value for which 
\begin{equation}
	\mathrm{Pr}[r_{\min}<r_{\min}^\gamma|\hat{\mathbf{G}}^\mathrm{PL}] \leq \gamma\%.
\end{equation}
This corresponds to the notion of
\emph{outage capacity}. Note that here we consider the outage in minimum rate 
over the entire network instead of over individual links. Instead of computing
the outage rate of each link, as in the earlier work \cite{junwang,dallanese} 
and the conference version of this paper \cite{icassp}, the percentile rate is
taken \emph{after} the network utility is computed. %\footnote{In \cite{icassp}, we didn't adopt the notion of robustness over the network level, due to the sum-rate objective we had implying collaborated transmission across the network, which is impractical for independent D2D links. In this paper however, the min-rate objective we are considering implies fairness and starvation-avoidance, which is suitable to be enforced under variations over the network level.}), i.e., over channel realizations, the percentile is taken \emph{after} the network utility is computed. 

The robust optimization problem for the D2D network can now be formulated as
follows. We assume that a central controller has access to only the path-loss
components and the beamforming gains of all the channels, and seeks to find
a set of \emph{robust} power allocations that work well over different 
realizations of the shadowing and fast fading components. 
This D2D wireless network power control problem is therefore 
one instance of the robust optimization problem $\mathcal{P}$ in
Section~\ref{sec:prob_II}. We have the following correspondence: % (where we omit the trivial correspondence between the power loading solutions and the optimization variables, as both are denoted by $\mathbf{x}$): 
\begin{align}
    \mathbf{p} \leftarrow& \: \mathbf{G} \nonumber \\
    \mathbf{q} \leftarrow& \: \hat{\mathbf{G}}^\mathrm{PL} \nonumber \\
    u_{\mathbf{p}}(\mathbf{x}) \leftarrow& \: r_{\min} \nonumber \\
    u^\gamma \leftarrow& \: r_{\min}^\gamma \nonumber
\end{align}

The \emph{robust power control} for minimum-rate maximization problem in D2D wireless networks, which we denote by $\mathcal{P}_{\text{D2D}}$, is then 
\begin{subequations}\label{robust_D2D_prob}
\begin{align}
\underset{\mathbf{x}}{\text{maximize}}\quad& r_{\min}^\gamma\\
\text{subject to}\quad& \mathrm{Pr}[r_{\min}<r_{\min}^\gamma | \hat{\mathbf{G}}^\mathrm{PL}]\leq\gamma\%\\
& r_{\min}\leq r_i,\;\forall i\\
& 0\leq x_i \leq 1,\;\forall i \\
&\text{and }(\ref{equ:D2D_uncertainty}), (\ref{equ:D2DRate}) .\nonumber
\end{align}
\end{subequations}

%%%%%%%%%%%%%%%%%%%%%%%%%%%%%%%%%%%
%%%%%%%%%%Method Start%%%%%%%%%%%%%
%%%%%%%%%%%%%%%%%%%%%%%%%%%%%%%%%%%

\section{Uncertainty Injection: A Deep Learning Training Scheme for Robust Optimization}\label{sec:method}

This section presents a novel \emph{uncertainty injection} scheme for training
deep learning models for solving robust optimization problems.  The goal is to
train a model to produce solutions that maximize a utility subject to an outage
constraint. The proposed training scheme is applicable to a wide variety of
problems and different neural network architectures, requiring only the
assumption that the utility can be easily evaluated under different problem
parameters. 

\subsection{Sample-based Uncertainty Distribution Characterization}\label{sec:method_I}

Natural phenomena often induce uncertainties that are difficult to
characterize accurately. Their distributions can have unbounded support, and
are often not easily expressed by tractable mathematical expressions.
Traditional robust optimization algorithms rely on building mathematical models
for the distributions of the uncertainties, then performing optimization based
on these models. Imposing mathematical models has two major drawbacks: the
models may not fully characterize the true uncertainty distributions; further
the subsequent optimizations are often not analytically tractable, may require
extra simplifications, and may result in significant computational complexities
for obtaining the final solution.

In this paper, we advocate characterizing parameter uncertainties via
\emph{sampling}. Samples of the uncertainty can be straightforward to generate
for many realistic applications. With sufficient number of samples, we can
estimate the statistical objective numerically, thus eliminating the need for
mathematical models and the associated analysis. Sampling has been widely used
in many machine learning algorithms, such as stochastic gradient descent
\cite{sgd}, evolutionary algorithm \cite{evolution}, variational inference 
\cite{vae}, and so on. In this paper, we use a sampling strategy to develop 
robust optimization algorithms that are more flexible and
more computationally efficient than traditional mathematical optimization 
algorithms.  

\subsection{Uncertainty Injection Scheme}\label{sec:method_II}

We propose an approach of training a neural network to solve the robust
optimization problem $\mathcal{P}$.  The neural network takes measurements
of the problem parameters as input, i.e., $\mathbf{q}$, and computes a robust 
solution $\mathbf{x}$. The mapping from $\mathbf{q}$ to $\mathbf{x}$ can be
modeled by any neural network structure. The key for our training scheme is the
injection of uncertainty realizations of the problem parameters into the 
training process, after $\mathbf{x}$ is computed.

Specifically, during the training stage, after the neural network computes the
solutions $\mathbf{x}$ based on $\mathbf{q}$, we inject a large number of $L$
problem parameters realizations $\{\mathbf{p}_1,\mathbf{p}_2,\mathbf{p}_3,\dots,
\mathbf{p}_L\}$, sampled from $f_{\mathbf{p}|\mathbf{q}}(\mathbf{p}|\mathbf{q})$,
at the output layer by 
computing the objective values of the optimization problem for each of
these samples of problem parameters, i.e.,
$u_{\mathbf{p}_1}(\mathbf{x}),u_{\mathbf{p}_2}(\mathbf{x}),\dots,u_{\mathbf{p}_L}(\mathbf{x})$.
From these set of objective values, we can obtain $\hat{u}^\gamma$,
an empirical estimate of $u^\gamma$, by ranking the $L$ objective values and
take the ${(L\gamma)}$-th lowest value (with linear interpolation if $L\gamma$ is not an integer). Based on the gradients
derived from $\hat{u}^\gamma$, we update the neural network weights $\Theta$ to
improve the robust objective performance in an unsupervised learning fashion. The idea is that during testing, the neural network can compute $\mathbf{x}$ 
based only on $\mathbf{q}$, but can nevertheless achieve statistical robustness 
against the unseen uncertainties in $\mathbf{p}$.

\begin{figure*}
    \centering
    \includegraphics[width=0.75\textwidth]{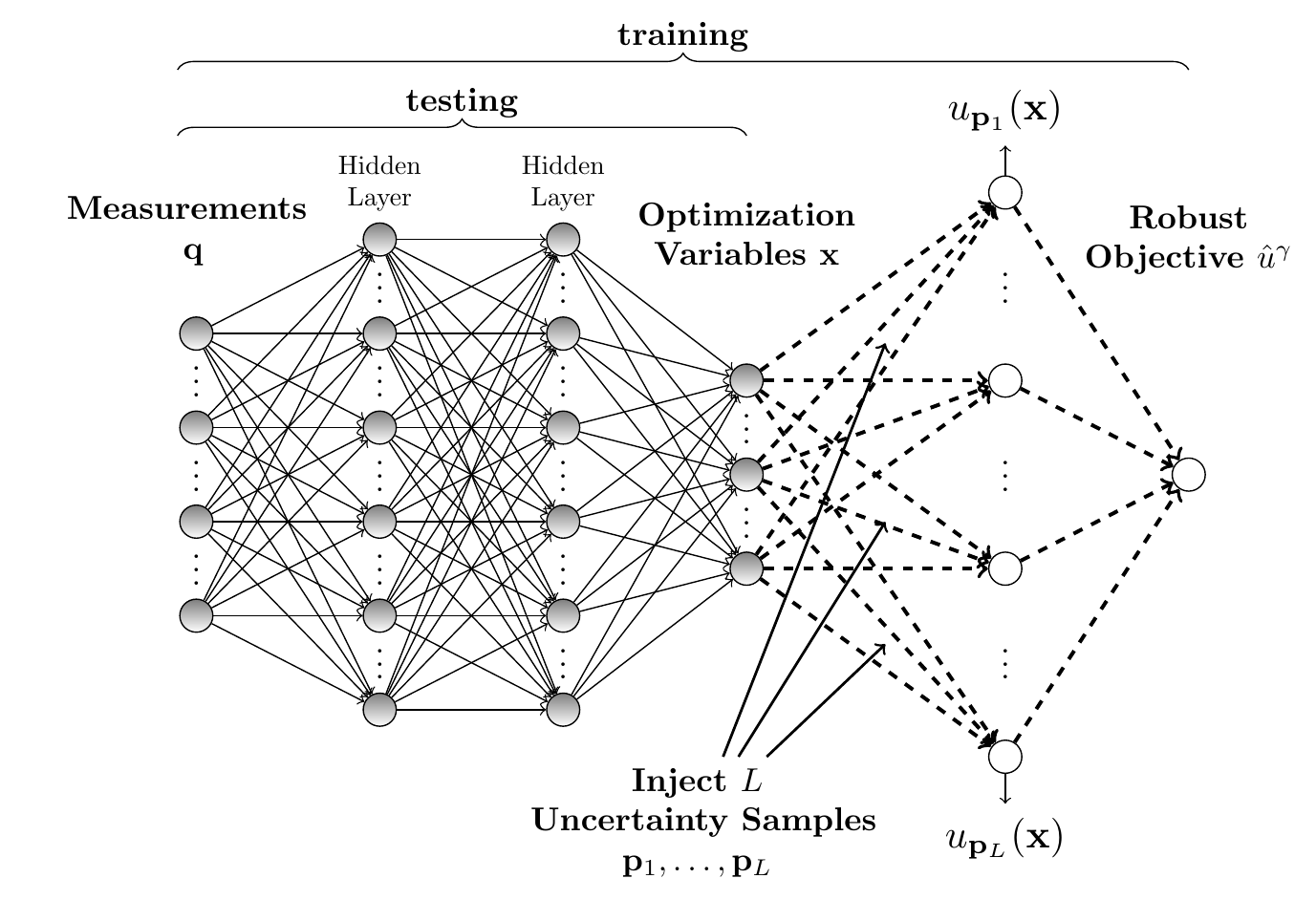}
    \caption{Uncertainty injection scheme for training a neural network for robust optimization}
    \label{fig:neuralnet}
\end{figure*}

The overall neural network structure is shown in Fig.~\ref{fig:neuralnet}. During training, the computation flows all the way towards the end for computing $\hat{u}^\gamma$; while during testing or for applications, the computation stops at the output layer for the optimization variables $\mathbf{x}$. 

We emphasize that the proposed method of training with uncertainty realizations
injection is fundamentally different from the idea of \emph{data augmentation}, in which various
transformations or noises are applied to the training data before feeding to
the model \cite{resnet, chenyu}. Our method is also fundamentally different
from existing literature on \emph{noise injection} for robust deep learning, in which
noises are injected to the inputs \cite{bishop,neuralsmithing,shaham,moosavi,madry},
the neural network parameters \cite{graves},
the activation functions \cite{benpoole,caglar},
and the gradients \cite{neelakantan,mozhou}, as already mentioned earlier. 
In contrast, the proposed strategy injects uncertainties \emph{after} the
neural network output layer, and updates the neural network model in an unsupervised fashion, based on an objective which is a stochastic function of the input (i.e., optimizing the $\gamma$-th percentile of $u_{\mathbf{p}}(\mathbf{x})$ when the neural network input provided is $\mathbf{q}$). This is a sensible strategy in the robust optimization setting because the task at hand is an explicitly formulated mathematical optimization problem, and the output of the neural network is a solution to this problem, for which 
we can readily evaluate its robustness under the uncertain problem parameters.
%together with an unsupervised learning scheme for maximizing the robust objective. 
%(i.e. deriving gradients directly from the empirical estimator $\hat{u}^\gamma$). 
The proposed strategy uses injected uncertainty realizations to compute an 
robust objective, then trains the neural network to maximize this objective. 
This training scheme allows the neural network to
implicitly learn the underlying uncertainty distribution for achieving
robustness in its solutions, which is impossible to do under data augmentation
or existing noise injection methods in the literature.

\subsection{Gradient Computation for Back-Propagation}\label{sec:method_III}

We now elaborate on the mathematical details of the gradient back-propagation
under the uncertainty injection training scheme. With $\mathbf{x}$ computed by
a usual neural network, the learned $\mathbf{x}$ is a differentiable function 
of the neural network parameters $\Theta$. In the meanwhile, %as stated in Section~\ref{sec:prob_II}, we require 
the objective function $u_{\mathbf{p}}(\mathbf{x})$ is assumed to be 
differentiable (almost everywhere) over $\mathbf{x}$ for fixed problem parameters $\mathbf{p}$. 
Lastly, the sample-based empirical percentile value $\hat{u}^\gamma$, even 
with possible linear interpolation between two points, is differentiable 
with respect to the sampled objective values $\{u_{\mathbf{p}_1}(\mathbf{x}),u_{\mathbf{p}_2}(\mathbf{x}),\dots,u_{\mathbf{p}_L}(\mathbf{x})\}$
in a \emph{local neighborhood}. This is because the percentile value does not
depend on most of the sampled objective values, except one or two closest to the
target percentile. When $L$ is finite, the rankings of the sampled objective
values closest to the target percentile do not change in the local neighborhood, so differentiability can be ensured locally almost surely.

We emphasize that the uncertainty injection scheme is not restricted to the
percentile function as the notion of robustness. As long as robustness is
defined by a function whose sample-based approximation is differentiable over
the samples, 
%(e.g. sample mean, sample variance, sample range, etc.), 
we can train a deep learning model for this robustness objective using the
uncertainty injection scheme. 

Taken together, we have that the empirical robust objective $\hat{u}^\gamma$ is
a differentiable function over the neural network model parameters $\Theta$.
Therefore, the neural network can be optimized with the stochastic gradient
descent method when being trained with the uncertainty injection scheme.

To compute the gradients, let $\mathbf{p}_i$ be the uncertainty realization
(among the set of $L$ injected uncertainty samples) whose corresponding
objective $u_{\mathbf{p}_i}(\mathbf{x})$ is the empirical $\gamma$-th 
percentile value of the set of sample objectives, i.e.,
$\hat{u}^\gamma$. The gradient of $\hat{u}^\gamma$ with respect to the neural
network parameters $\Theta$ can be obtained by the chain-rule of
differentiation\footnote{If linear interpolation is needed for
the empirical percentile, the expression for the
gradients would involve a linear combination of two of realizations.}:
\begin{align}\label{equ:gradient_sample}
	\frac{\partial \hat{u}^\gamma}{\partial \Theta} = 
        \frac{\partial \hat{u}^\gamma}{\partial u_{\mathbf{p}_i}(\mathbf{x})}
        \frac{\partial u_{\mathbf{p}_i}(\mathbf{x})}{\partial \mathbf{x}}
        \frac{\partial \mathbf{x}}{\partial \Theta} 
	= 
        \frac{\partial u_{\mathbf{p}_i}(\mathbf{x})}{\partial \mathbf{x}}
        \frac{\partial \mathbf{x}}{\partial \Theta},
\end{align}
where the term\footnote{Or, it is a linear combination of two constants if interpolation is needed to compute the empirical percentile.} 
$\frac{\partial \hat{u}^\gamma}{\partial u_{\mathbf{p}_i}(\mathbf{x})}$ is 1;
the term $\frac{\partial u_{\mathbf{p}_i}(\mathbf{x})}{\partial\mathbf{x}}$ can 
be computed\footnote{At points of non-differentiability, we can take a supergradient.} based on the uncertainty realization $\mathbf{p}_i$; and lastly, 
the term $\frac{\partial \mathbf{x}}{\partial \Theta}$ depends on the number of 
activation layers and the values of $\Theta$ themselves in the neural network. 
In other words, the gradient of $\gamma$-th percentile robust objective is just the gradient of the objective corresponding of the $\gamma$-th percentile realization $\mathbf{p_i}$.

While the exact expressions for these gradients depend on the neural network structure and the objective function $u_{\mathbf{p}}(\mathbf{x})$, %and the mathematical definition of the robust objective. Nonetheless, 
all of these differentiations are readily computed by popular deep learning frameworks such as Pytorch \cite{pytorch} or Tensorflow \cite{tensorflow}. Therefore, the uncertainty injection scheme adds very little complexity to the training of deep neural networks, and no extra complexity at testing. %thus achieving great algorithmic efficiency.

\subsection{Sample-Based Gradient Estimator}\label{sec:method_IV}
The previous section establishes the gradient computation for optimizing the robust objective and shows that the proposed robust optimization process can be easily implemented in the usual deep learning computation frameworks. We now show that the sample-based gradient in (\ref{equ:gradient_sample}) is an asymptotically unbiased estimator of the true desired gradient.

Let $u^\gamma$ be the true $\gamma$-th percentile of $u^{\mathbf{p}}(\mathbf{x})$
under $\mathbf{p} \sim f_{\mathbf{p}|\mathbf{q}}(\mathbf{p}|\mathbf{q})$, 
i.e., the \emph{population statistic}. %; meanwhile, following the notion in Section~\ref{sec:method_III}, 
Let $\hat{u}^\gamma$ be the $\gamma$-th percentile of the samples $\{u_{\mathbf{p}_1}(\mathbf{x}), u_{\mathbf{p}_2}(\mathbf{x}), \dots\,u_{\mathbf{p}_L}(\mathbf{x})\}$, i.e.
the \emph{sample statistic}. Let $\frac{\partial u^\gamma}{\partial \Theta}$ be the true gradient for updating the neural network parameters assuming that the gradient exists, (or a supergradient if the gradient does not exist).

We now examine the sample-based gradient in (\ref{equ:gradient_sample}) under
the expectation. As described in Section~\ref{sec:method_II} and
Section~\ref{sec:method_III}, the sample statistic $\hat{u}^\gamma$ is obtained
by sorting the sample objective values then finding the index corresponding to 
the percentile in the samples, (i.e., with sample size being $L$, we find the
value ranked the $(L\gamma)$-th lowest, with linear interpolation if necessary).
As shown in \cite{vanzwet,chatillon}, under fairly general
conditions, this sample statistic is an asymptotically unbiased estimator to
the true population statistic $u^\gamma$. Therefore, we have: 
\begin{align}
    \lim_{L\to\infty}\mathbb{E}[\hat{u}^\gamma] = u^\gamma .
\end{align}
It can be seen that
the expectation of the gradient (\ref{equ:gradient_sample}), in the asymptotic limit of large sample size, is just
    $\frac{\partial u^\gamma}{\partial \Theta}$:
\begin{align}
    \lim_{L\to\infty}\mathbb{E}\left[\frac{\partial\hat{u}^\gamma}{\partial \Theta}\right] 
    =&\: \frac{\partial\lim_{L\to\infty}\mathbb{E}[\hat{u}^\gamma]}{\partial \Theta} \\ \label{equ:expectation_limit}
    =&\: \frac{\partial u^\gamma}{\partial \Theta}
\end{align}
where (\ref{equ:expectation_limit}) follows from the linearity of both the
limit and the expectation operators, and the fact that 
%assuming regularity conditions that allow 
the interchange of the limit, the expectation and the derivatives is allowed
under certain regularity conditions. Assuming
that such regularity conditions hold for the distribution
$u_{\mathbf{p}}(\mathbf{x})$, we have that in the uncertainty injection scheme,
the gradients used in updating the neural network weights are asymptotically
unbiased estimators of the true gradients, which is a desirable property for
the convergence of the training of deep learning models. We note that this
analysis resembles the proof that the gradient estimator of the stochastic
gradient descent (SGD) algorithm \cite{sgd} is unbiased.

%(except that in SGD, the gradient estimator is strictly unbiased).

\section{Applications of Uncertainty Injection in Wireless Communications}
\label{sec:app_wireless}

In this section, we describe the application of the uncertainty injection
training scheme for solving robust optimization in wireless communication
problems as described in Section~\ref{sec:prob_wireless}.
Further, we describe the benchmarks that the proposed training method would
be compared with in the numerical evaluation.

\subsection{Neural Network Architecture and Uncertainty Injection}\label{sec:exp_I}

For the two applications as described 
in Section~\ref{sec:prob_III} and Section~\ref{sec:prob_IV}, 
given the inputs as the estimated MIMO wireless
channels or the path-loss components in a D2D wireless network, the neural
network computes robust maximum minimum-rate solutions for power loading at 
the base station or power control among D2D links.

Robustness is achieved using the uncertainty injection training scheme, %as elaborated in Section~\ref{sec:method_II}, 
in which a large set of wireless channel realizations are sampled and 
injected at the output of the neural network. A set of minimum rates 
can then be computed, with one minimum rate for each of the channel
realization. This set of rates provides an empirical distribution of
the minimum rate over the uncertain channels. We then take the
$\gamma$-th percentile value of this set of rates $\hat{r}_{\min}^\gamma$ as
the empirical estimation for $r_{\min}^\gamma$. Through computing gradient and
performing gradient ascent on $\hat{r}_{\min}^\gamma$, %which is a function of neural network parameters $\Theta$, 
the neural network is trained towards the direction of improving the robust
minimum rate. 

To explore the full potential of the training scheme with deep learning, we
design the neural network based on the most general architecture: the fully connected neural networks. Following an
input layer which takes channel measurement values $\mathbf{q}$, we adopt 
fully connected
hidden layers each with \emph{ReLu} non-linearity activations. In the output
layer that computes the solutions for optimization variables $\mathbf{x}$,
proper non-linear activations are used to account for the constraints 
on the variable $\mathbf{x}$.

During the training for each single input, we sample and inject $L=1000$ uncertainty realizations. Correspondingly, at any point of training, following each neural
network forward path, we compute 1000 $u_{\mathbf{p}}(\mathbf{x})$ objective values 
based on the $\mathbf{x}$ computed by the neural network. The $\gamma$-th percentile objective among these 1000 $u_{\mathbf{p}}(\mathbf{x})$ objective values is then selected for computing the
gradient and performing gradient ascent on the neural network parameters $\Theta$. 

\subsection{Robust Objective vs. Nominal Objective}\label{sec:exp_II}

We show two types of objectives in the numerical simulations: the robust objective and the nominal objective. The robust objective is as specified by the robust optimization formulation $\mathcal{P}$, i.e., the $\gamma$ percentile value of the objective distribution achieved by the solution $\mathbf{x}$, as induced by the uncertainty distribution. 

On the other hand, the \emph{nominal objective} is the objective value achieved by the solution $\mathbf{x}$, under the measured (or estimated) input values as if they are the true channel parameter values. Under our notation, the nominal objective is denoted as $u_{\mathbf{q}}(\mathbf{x})$.

We show the results for the nominal objective, because it is the objective that
deterministic optimization algorithms (i.e. the non-robust optimization
algorithms) would optimize, assuming that the measured or estimated problem
parameters are entirely accurate.  The relative performances of all the methods
in terms of their nominal objective and the robust objective reveal the
effectiveness of actively accounting for the channel uncertainties during
optimization.

\subsection{Deep Learning without Uncertainty Injection}\label{sec:exp_III}

To fully illustrate the benefits of the proposed the training scheme, we include a benchmark method where an identical neural network is trained with the same dataset and hyper-parameters, \emph{but without uncertainty injection}. Specifically, we train the neural network with the same unsupervised learning procedure, except that we do not implement any uncertainty injection. 
Instead, during training the gradient is derived from the nominal objective $u_{\mathbf{q}}(\mathbf{x})$, computed directly based on the measured or estimated channel parameters $\mathbf{q}$.

We note that to the best of the authors knowledge, there are no efficient
analytic mathematical optimization methods that can account for channel
uncertainties in these problem settings.

%%%%%%%%%%%%%%%%%%%%%%%%%%%%%%%%%%%
%%%%%%Simulation Start%%%%%%%%%%%%%
%%%%%%%%%%%%%%%%%%%%%%%%%%%%%%%%%%%
\section{Experimental Validation}\label{sec:exp}
\newcommand{\minitab}[1]{\begin{tabular}{@{}c@{}}#1\end{tabular}}

This section provides numerical results for the proposed uncertainty injection scheme on two wireless network applications, namely, the robust power loading problem for the multiuser MIMO downlink, and the robust power control problem for D2D wireless networks.

\subsection{Multiuser MIMO Downlink Environment}\label{sec:exp_IV}

%This subsection describes the MIMO wireless networks settings. We set the number of antennas equipped at the base station as $M=4$. We start with a total number of 5 users present within the network, and then perform a primitive user scheduling heuristic and set one user to be inactive. Specifically, based on estimated channels $\hat{\mathbf{H}}$ (as in Section~\ref{sec:prob_IV}), we find two users, namely user $k$ and user $l$, with the most similar channels $\hat{\mathbf{h}}_k$ and $\hat{\mathbf{h}}_l$ in terms of directions, and then remove the user with weaker channels (lower channel vector norm) from the transmission list. After the user scheduling, we are left with $K=4$ users that the base station transmits to. We assume the i.i.d. Rayleigh fading model for MIMO channels $\mathbf{H}$, with each channel following a circular symmetric complex Gaussian distribution $h_{kj}\sim\mathcal{CN}(0,1),\:\forall k,j$.
Consider a base station equipped with $M=4$ antennas and serving $K=4$ users. 
We assume the i.i.d. Rayleigh fading model for the MIMO channel $\mathbf{H}$,
with each channel entry following a circularly symmetric complex Gaussian
distribution $h_{kj}\sim\mathcal{CN}(0,1),\:\forall k,j$.
%As mentioned in Section~\ref{sec:prob_IV}, before the transmission, each user sends a pilot sequence to the base station for channel estimations. For this simulation, we assume that every user only transmits one pilot symbol at the power of 0.19W and a background noise of 0.01W (both at the base station when receiving pilots and at users when receiving data). According to the MMSE estimation \cite{mmse_channel}, 
We assume that the channel estimation error $e_{kj}$ as in 
(\ref{equ:MIMO_uncertainty}) has a variance of $\sigma_e^2 = 0.075$. During
data transmission, we assume the total power constraint $P$ of 1W over a
bandwidth of 10MHz, and an effective background noise level of -75dBm/Hz 
(e.g., after accounting for pathloss and out-of-cell interference).

Prior to the power loading optimization, based on the estimated channels
$\hat{\mathbf{H}}$ using minimum mean-square estimation (MMSE)
\cite{mmse_channel}, we utilize the RZF 
beamformers, which have a better performance than the ZF beamformers.
Specifically, we compute the unnormalized RZF beamformers as follows: 
\begin{align}
    \mathbf{B}' = \hat{\mathbf{H}}(\hat{\mathbf{H}}^H\hat{\mathbf{H}}+\alpha \mathbf{I})^{-1}
\end{align}
To choose an appropriate value for $\alpha$, we make the following observation
(which is verified by numerical simulations): as $\alpha$ approaches zero, 
the beamformers become more aggressive on nulling the interference, but also
become more sensitive with respect to channel uncertainty. As a
result, the nominal objective (as defined in Section~\ref{sec:exp_II}) 
increases while the robust objective degrades. To achieve a reasonable 
trade-off on the regularization factor, we select the value of $\alpha$ 
that results in the highest medium $r_{\min}$ under the channel
uncertainty distribution (assuming equal power allocation across the 
users for simplicity). This leads to a value of $\alpha=0.2$ in our setting. 

We then normalize the
beamformers across each column of $\mathbf{B}'$, i.e., the beamformers for each
user, to unit norm, to obtain the normalized beamformers $\mathbf{B}$. 

For the inputs to the neural network, we compute the \emph{effective channels} $\hat{\mathbf{H}}_{\text{eff}}$ based on $\hat{\mathbf{H}}$ and $\mathbf{B}$ as:
\begin{align}
    \hat{\mathbf{H}}_{\text{eff}} = \hat{\mathbf{H}}^H\mathbf{B}
\end{align}
and flatten the resultant matrix into a length-$(K\times K)$ vector. We note that this information is the measurement of the channel, i.e., playing the role of $\mathbf{q}$ in the robust optimization problem $\mathcal{P}$ as in Section~\ref{sec:prob_II}. %since no actual knowledge about the real channel $\mathbf{H}$ is required within the computation.

Provided with this input, the fully-connected neural network then computes the output $\mathbf{x}\in[0,1]^K$ as the power loading solution. We use the \emph{softmax} activation at the final output layer to enforce the power constraint $\sum_{i=1}^K x_i \le 1$. %Corresponding to the range of each $x_k$ as the percentage of power allocated for the user $k$, we use the element-wise sigmoid non-linearity activations at the output layer of the neural network.

Lastly, to train the neural network with uncertainty injection scheme for robust minimum-rate optimization, we randomly generate 1000 channel realizations of $\mathbf{H}$ according to (\ref{equ:MIMO_uncertainty}) and inject them into the neural network training flow. With a specific solution $\mathbf{x}$ computed by the neural network, we compute then take the minimum of the $K$ user rates achieved for each of the 1000 channel realizations. We then take the $\gamma$-th percentile of those 1000 minimum rates to obtain the empirical estimation $\hat{u}^\gamma$. 
The gradient to update the neural network weights is computed based on the channel realization corresponding to $\hat{u}^\gamma$.

During training, we use a minibatch size of 1000 distinct MIMO networks, with each training epoch including 50 minibatches. We train for a total of 500 epochs, with early stopping based on validation at the end of each epoch. We note that instead of having a fixed training set, we continuously generate new MIMO network wireless channels and their estimations, which is easy to do because of the readily available Rayleigh fading channel model. 

\begin{table*}
\caption{Multiuser MIMO Downlink 5-Percentile Robust Minimum-Rate Performance}
\centering
\begin{tabular}{c|cc}
\hline
Methods & Nominal Minimum Rate & Robust Minimum Rate  \\
\hline
\bigstrut Geometric Programming & 40.18Mbps & 5.41Mbps \\
\bigstrut \minitab{Deep Learning \textbf{without} \\ Uncertainty Injection} & 39.51Mbps & 6.52Mbps \\
\bigstrut \minitab{Deep Learning \textbf{with} \\ Uncertainty Injection} & 37.40Mbps & \textbf{8.03Mbps} \\
\bigstrut Uniform Power & 29.70Mbps & 5.88Mbps\\
%\bigstrut Random Power & 10.78Mbps & 2.71Mbps \\
\hline
\end{tabular}
\label{tab:mimo_minrates}
\end{table*}

\subsection{Multiuser MIMO Downlink Results}\label{sec:exp_V}
We adopt a structure for the neural network with 4 fully-connected layers,
each with 200 hidden neurons and the \emph{ReLU} non-linearity, except for the
output layer with the \emph{softmax} non-linearity to ensure that $\sum x_i=1$. 
(In the low-interference regime after RZF, the optimized $\mathbf{x}$ is likely
to be a one that uses up all the total power, i.e., (\ref{equ:robust_MIMO_prob_I}) is satisfied with equality.)
%producing $\mathbf{x}$, with $K=4$ output units and the \emph{Sigmoid} non-linearity (to cater to the power loading solution for 4 users, with the range as $[0,1]^4$). 
This same structure is also use in the deep learning without uncertainty
injection benchmark as described in Section~\ref{sec:exp_III}.

%For comparisons, we include four benchmarks, with the first two benchmarks (\emph{deep learning with regular training} and \emph{geometric programming}) specifically illustrating the importance of implicitly accounting for uncertainties in the optimization process:
%\begin{itemize}
%    \item \emph{Deep Learning with Regular Training}, as elaborated in Section~\ref{sec:exp_III}.
%    \item \emph{Geometric Programming}, providing the global-optimal solution with environmental parameters being deterministic.
%    \item \emph{Full Power}, allocating 100\% power for each user.
%%    \item \emph{Random Power}, allocating a percentage of power for each user uniformly selected from 0\%$\sim$100\%.
%\end{itemize}

We also compare with a geometric programming benchmark, which finds the 
global optimal solution for the minimum-rate maximization problem by 
representing the inverse of the signal-to-inference-and-noise ratio (SINR)
terms as posynomials \cite{mung}. Note that the formulation required by
geometric programming cannot incorporate channel uncertainties, because the
statistical robustness of the objective cannot be expressed by posynomials.  

\begin{figure}
    \centering
    \ifOneColumn
      \includegraphics[width=0.7\textwidth]{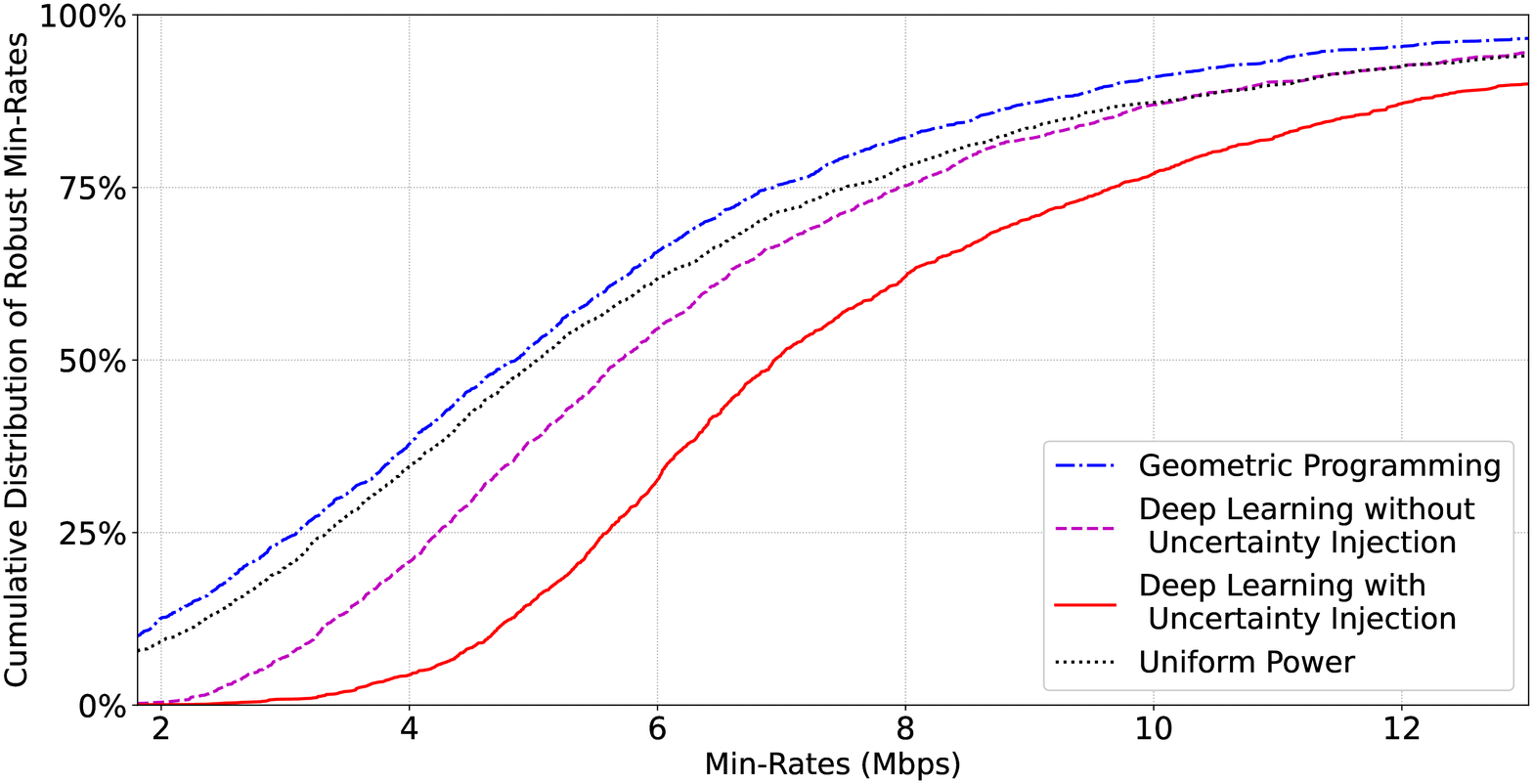}
    \else
      \includegraphics[width=9cm]{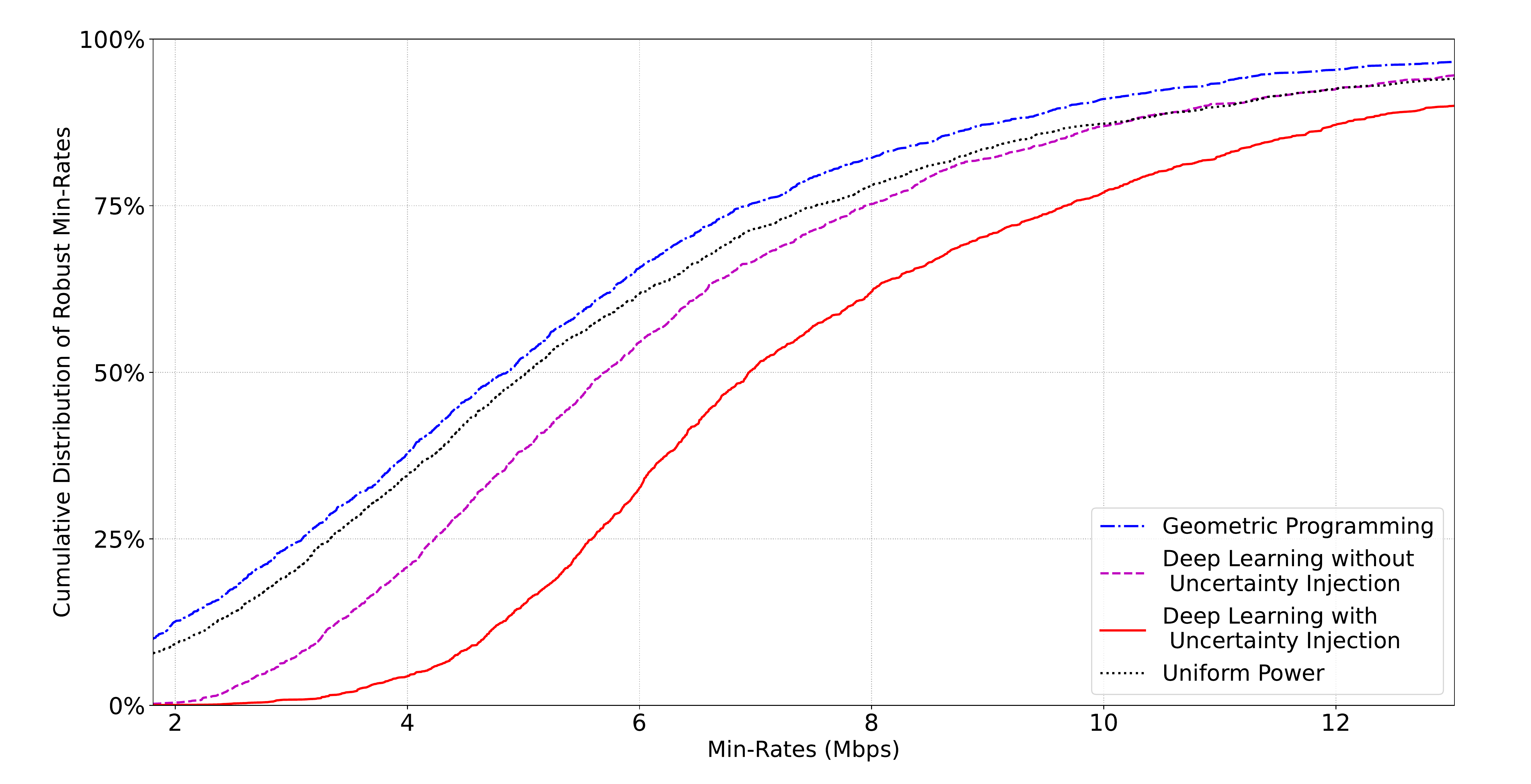}
    \fi
    \caption{Cumulative distribution of 5th-percentile robust minimum-rates in MIMO wireless networks over 2000 test channel layouts.}
    \label{fig:MIMO_robust_minrates_CDF}
\end{figure}

In the simulations, we set $\gamma=5$, i.e., we use the robust minimum 
rate at the 5th-percentile across the distribution induced by the channel 
uncertainties as the robust objective. We evaluate the robust minimum rate 
and the nominal minimum rate performances over 2000 testing layouts. For
evaluation of the robust objective, we also 
use 1000 randomly sampled channel realizations on each of the 2000 testing 
layouts to empirically estimate the 5th-percentile value of the minimum rates. 
The results are presented in Table~\ref{tab:mimo_minrates}. Furthermore, 
the cumulative distribution function (CDF) of these robust 5th-percentile minimum-rates
over all 2000 testing layouts are shown in Fig.~\ref{fig:MIMO_robust_minrates_CDF}. 

As the numerical results show, the neural network trained with the uncertainty
injection scheme achieves 23\% improvement on the robust minimum rate
over the same neural network trained without uncertainty injection. 
This shows
that the uncertainty injection scheme indeed encourages the deep learning 
model to learn and to produce solutions that are significantly more robust 
against the channel uncertainties than the state-of-the-art algorithms. %, which cannot be learned with regular deep learning training. 

Interestingly, the performance improvement over geometric programming is even
larger, at 48\% for the robust 5th-percentile minimum-rate. This is because 
the classical mathematical programming techniques exploit the particular
channel realization to the fullest extent in order to maximize the minimum-rate
objective, as shown in the nominal minimum rate result, but it does not account
for robustness across the channel uncertainties at all. Even training a neural
network to approximate the classical mathematical programming without
uncertainty injection would already achieve some robustness, as shown in
Fig.~\ref{fig:MIMO_robust_minrates_CDF}.  Significantly better robustness 
is achieved with training a deep neural network with uncertainty injection.

\subsection{D2D Wireless Network Environment}\label{sec:exp_VI}

For the D2D wireless network, we consider a number of D2D links randomly
deployed within a confined region, with the transceiver distances following
uniform distributions. We impose a minimum of 5-meter distance between any
transmitter and any receiver that do not belong to the same link. For the
path-loss, we follow the short-range outdoor model ITU-1411, with 5MHz
bandwidth at the carrier frequency of 25GHz (the mmWave range). 

\begin{figure}
    \centering
    \ifOneColumn
      \includegraphics[width=0.7\textwidth]{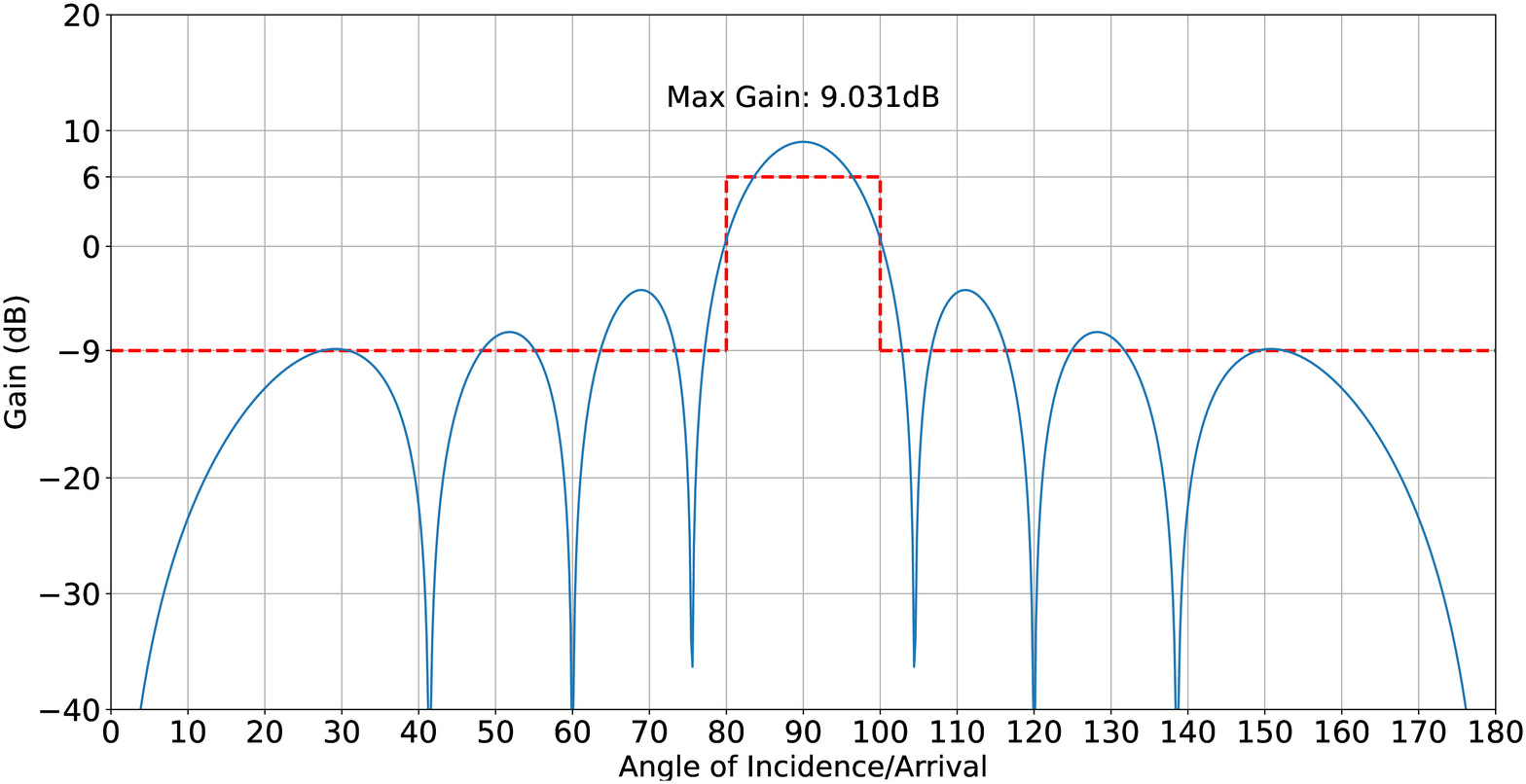}
    \else
      \includegraphics[width=9cm]{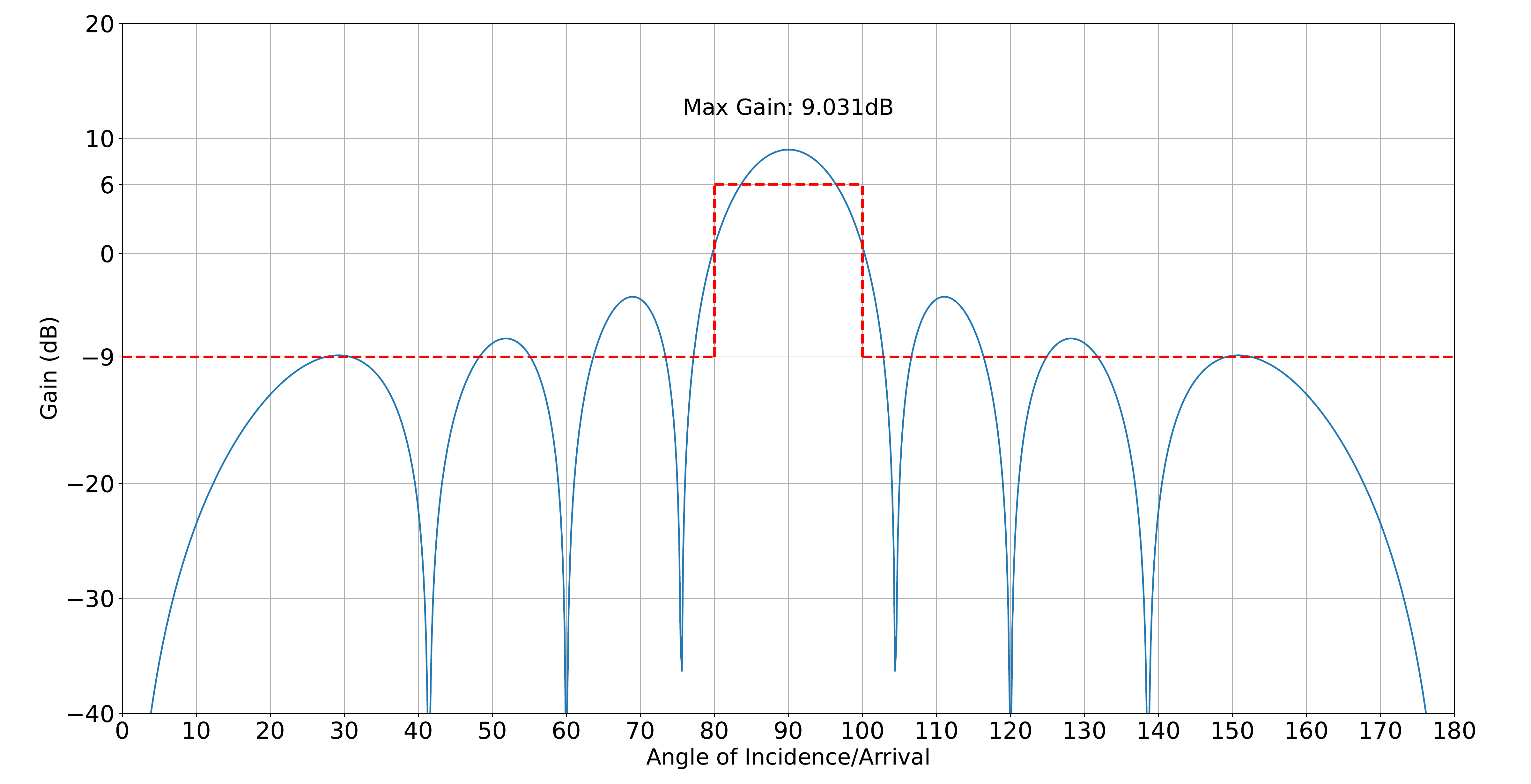}
    \fi
    \caption{D2D beamforming pattern at the transmitters and the receivers, and its approximation used in simulations.}
    \label{fig:D2D_beamformer}
\end{figure}

Every transmitter and receiver is equipped with a linear antenna array with 
$M=8$ antennas. %, all located at 1.5-meter height. 
%We employ the straight-forward directional beam-forming towards the direction towards the front perpendicular to the line the antennas array aligns. 
We use a beamforming pattern corresponding to a uniform linear array directly
aimed between the transmitter and the receiver.
The beamforming gains are approximated as follows: 9dB gain for the
direct links (as the transmitter and receiver pair can align with each other
precisely), 6dB gain at the main lobe direction
with an angle from $-10$ to $+10$ degrees, and $-9$dB gain at the side lobe directions.
In Fig.~\ref{fig:D2D_beamformer}, the corresponding beamforming gain pattern
and its approximation are plotted for $M=8$. At each transmitter, we
assume a maximum transmit power of 30dBm. We also assume a background noise
level of $-169$dBm/Hz.

We incorporate the following channel uncertainty models for each direct and interfering links: 
\begin{itemize}
    \item Shadowing with log-normal distribution with 8dB standard deviation;
    \item Rayleigh fading with i.i.d. circularly
	    symmetric complex Gaussian distribution with unit variance.
\end{itemize}
Furthermore, to illustrate that the training process can be applied to different scenarios, we test the robust performances in three different settings with 
1000 wireless networks layouts under each setting, as in Table \ref{tab:settings}. For each layout, we sample 1000 channel realizations during testing to obtain an empirical approximation of the robust minimum rate across the entire network. 

\begin{table}
\caption{D2D Wireless Environment}
\centering
\begin{tabular}{|c||c|c|c|}
\hline
Setting & \minitab{Number of \\ Links ($N$)} & \minitab{Region Area \\ $(m^2)$} & \minitab{Direct-Link \\ Distance Distribution} \\
\hline
{A} & 10 & $150\times150$ & 5m$\sim$15m \\
\hline
{B} & 10 & $200\times200$ & 20m$\sim$30m \\
\hline
{C} & 15 & $300\times300$ & 10m$\sim$30m \\
\hline
\end{tabular}
\label{tab:settings}
\end{table}

\subsection{D2D Wireless Networks Results}\label{sec:exp_VII}

We adopt a neural network structure with 5 fully-connected layers, each with
6$N^2$ hidden neurons and the \emph{ReLU} non-linearity, except for the output
layer, which produces $\mathbf{x}$ with $N$ output units and has a \emph{sigmoid}
non-linearity (to ensure that the power control solution for the $N$ links has
a range of $[0,1]^N$). This same structure is also used for the benchmark of
deep learning without uncertainty injection as described in
Section~\ref{sec:exp_III}.

Because of the path-loss, the direct and the interfering links are often of
different orders of magnitudes, making it difficult to obtain meaningful
updates at the beginning of training. To make training
effective, we adopt \emph{input normalization} on the input path-loss values
(for both training and testing), where each of the $N^2$ inputs of the 
path-loss values are normalized independently based on the statistics 
computed from the training set.

\begin{figure}
    \centering
    \ifOneColumn
      \includegraphics[width=0.7\textwidth]{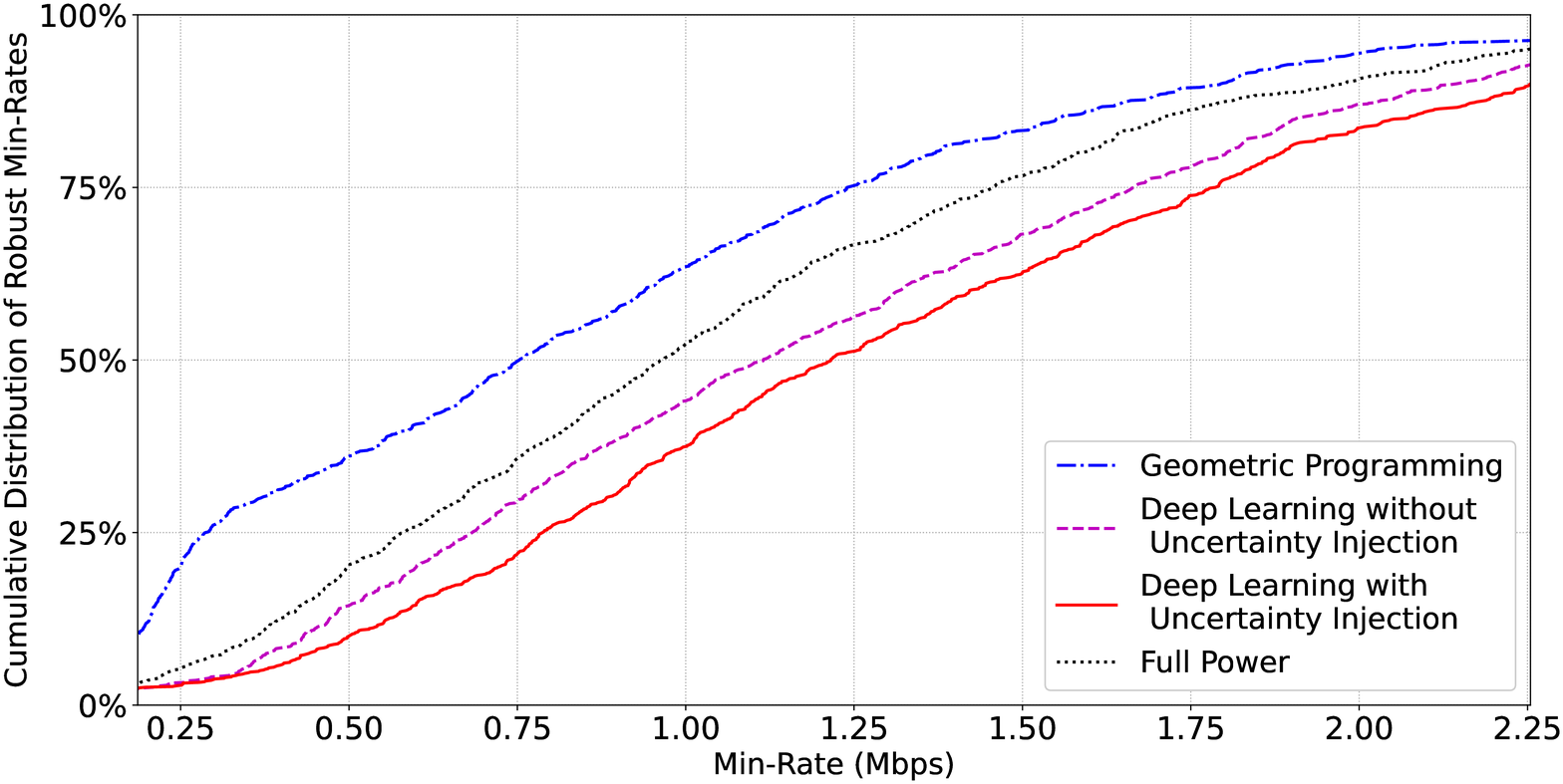}
    \else
      \includegraphics[width=9cm]{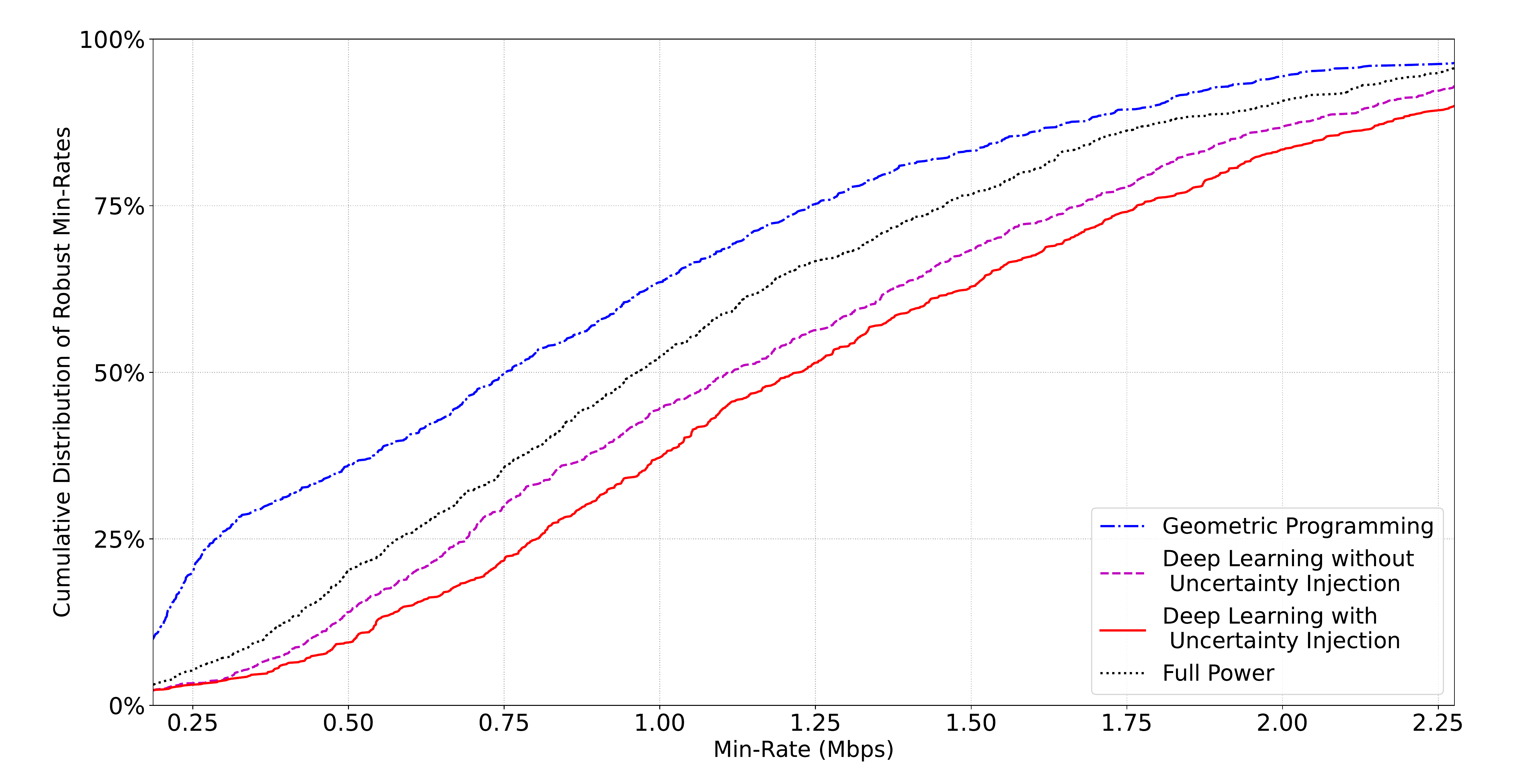}
    \fi
    \caption{Cumulative distribution of robust minimum-rates in D2D wireless networks under test setting B.}
    \label{fig:D2D_robust_minrates_CDF}
\end{figure}

For the D2D network simulations, we set $\gamma=10$, i.e., we use
the minimum rate at the 10th percentile across the distribution induced by the
channel uncertainties as the robust objective. We present test results on the
robust minimum-rate
obtained by the neural network based only on the path-loss and the beamforming 
gain values as the inputs. Similar to Section~\ref{sec:exp_V},
we include two benchmarks for illustrating the importance of robust
optimization: deep learning without uncertainty injection, and the geometric 
programming (which can provide globally optimal solutions if the channel
uncertainties are not considered).
Results are shown in Table \ref{tab:D2D_minrates}.
Fig.~\ref{fig:D2D_robust_minrates_CDF} presents CDF curves of the robust
minimum-rates achieved by all methods % (where we use \emph{Robust Deep Learning} and \emph{Regular Deep Learning} to denote deep learning with uncertainty injection and with regular training respectively) 
over 1000 testing wireless networks, generated under test setting B in Table II. 

\begin{table*}[ht!]
\caption{D2D Wireless Networks Robust Minimum-Rate Performance}
\centering
\begin{tabular}{c|cccccc}
\hline
\multirow{2}{*}{Methods} & \multicolumn{2}{c}{Setting A} & \multicolumn{2}{c}{Setting B} & \multicolumn{2}{c}{Setting C}\\
& Nominal & Robust & Nominal & Robust & Nominal & Robust\\
\hline
\bigstrut Geometric Programming & 46.84Mbps & 2.72Mbps & 37.05Mbps & 0.86Mbps & 40.36Mbps & 0.85Mbps \\
\bigstrut \minitab{Deep Learning \textbf{without} \\ Uncertainty Injection} & 44.86Mbps & 3.78Mbps & 35.41Mbps & 1.21Mbps & 37.56Mbps & 1.37Mbps \\
\bigstrut \minitab{Deep Learning \textbf{with} \\ Uncertainty Injection} & 43.18Mbps & \textbf{4.03Mbps} & 33.81Mbps & \textbf{1.31Mbps} & 35.70Mbps & \textbf{1.41Mbps} \\
\bigstrut Full Power & 37.43Mbps & 3.29Mbps & 28.61Mbps & 1.07Mbps & 30.47Mbps & 1.13Mbps \\
%\bigstrut Random Power & 30.91Mbps & 2.13Mbps & 22.26Mbps & 0.66Mbps & 23.60Mbps & 0.67Mbps \\
\hline
\end{tabular}
\label{tab:D2D_minrates}
\end{table*}

As shown in these results, under various interference levels, link densities,
and transceiver distance distributions, the neural network trained with
uncertainty injection
consistently produces more robust solutions. The performance gain due to the
proposed uncertainty injection training is up to 9\%, 
as compared to a deep neural network without uncertainty injection.
%We note that the margins being relatively small compared to Section~\ref{sec:exp_V}. We note that much larger margins by our method (over 10\%) have been observed from our extensive simulations, under D2D network settings with higher interference levels.  These observations indicate that with robust optimization accounting for unfavorable uncertainty realizations, larger margins on the robust minimum-rates can be achieved when the interference levels are higher. This fits the intuition as the robustness, i.e., the worst-case performances, are mostly limited by unusually large interferences from undesirable channel realizations, which are more prominent when the general interference level is high. We omitted presenting those results as the SINRs under such interference levels are too low for the modern transmission standards. 
%Nonetheless, regardless the SINR regime, the improvements with uncertainty injection training are shown to be evident and consistent given the variety of experimental settings, as well as the large number of samples generated for obtaining the numerical results. 
The gain is much higher, in the range of 48-66\%, when compared with the geometric programming benchmark, which is a globally optimal solution that does not account for channel uncertainty.
In fact, geometric programming does not even perform as well as simply using full power in term of the robust minimum rate, although as expected it does much better in term of nominal minimum rate.

%%%%%%%%%%%%%%%%%%%%%%%%%%%%%%%%%%%%
%%%%%%%%%%%%Conclusion Start%%%%%%%%
%%%%%%%%%%%%%%%%%%%%%%%%%%%%%%%%%%%%
\section{Conclusion}\label{sec:conclusion}

This paper proposes a novel uncertainty injection training method for deep
learning models to perform robust optimization against parameter
uncertainties. We consider a robust optimization problem that aims to
compute robust solutions based only on the measured or the estimated problem
parameters, while can still perform well with parameter
uncertainties. Traditional robust optimization techniques addressing such
problems require complicated mathematical characterizations of the parameter
uncertainties, which are often difficult to analyze and may not reflect
realistic environments accurately. Instead, sample-based uncertainty modeling
is an attractive alternative because of its simplicity and accuracy. 

The proposed training scheme utilizes the capability of deep neural networks
to learn from samples of uncertainties. By injecting uncertainties, an 
empirical estimate of the robust objective can be obtained and used
for updating the neural network model parameter during training. 
The proposed uncertainty injection scheme are applicable to a variety of 
robust optimization problems, while having low computational complexity.
To illustrate the effectiveness of this method, we present promising numerical 
simulation results on two wireless communication applications.
We believe that the proposed method opens up an avenue for wider 
application of deep learning based robust optimization under realistic scenarios, where 
uncertainties are ubiquitous and highly agnostic.

\bibliographystyle{IEEEtran}
\bibliography{citations}

\begin{thebibliography}{10}
\providecommand{\url}[1]{#1}
\csname url@samestyle\endcsname
\providecommand{\newblock}{\relax}
\providecommand{\bibinfo}[2]{#2}
\providecommand{\BIBentrySTDinterwordspacing}{\spaceskip=0pt\relax}
\providecommand{\BIBentryALTinterwordstretchfactor}{4}
\providecommand{\BIBentryALTinterwordspacing}{\spaceskip=\fontdimen2\font plus
\BIBentryALTinterwordstretchfactor\fontdimen3\font minus
  \fontdimen4\font\relax}
\providecommand{\BIBforeignlanguage}[2]{{%
\expandafter\ifx\csname l@#1\endcsname\relax
\typeout{** WARNING: IEEEtran.bst: No hyphenation pattern has been}%
\typeout{** loaded for the language `#1'. Using the pattern for}%
\typeout{** the default language instead.}%
\else
\language=\csname l@#1\endcsname
\fi
#2}}
\providecommand{\BIBdecl}{\relax}
\BIBdecl

\bibitem{icassp}
W.~Cui, K.~Shen, and W.~Yu, ``Deep learning for robust power control for
  wireless networks,'' in \emph{IEEE Int. Conf. Acoust. Speech Signal Process.
  (ICASSP)}, May 2020.

\bibitem{alexnet}
A.~Krizhevsky, I.~Sutskever, and G.~E. Hinton, ``Imagenet classification with
  deep convolutional neural networks,'' in \emph{Adv. Neural Inf. Process.
  Syst. (NIPS)}, Dec. 2012.

\bibitem{maas}
A.~L. Maas, R.~E. Daly, P.~T. Pham, D.~Huang, A.~Y. Ng, and C.~Potts,
  ``Learning word vectors for sentiment analysis,'' in \emph{Annu. Meeting
  Assoc. Comput. Linguistics (ACL)}, Jun. 2011, pp. 142--150.

\bibitem{resnet}
K.~He, X.~Zhang, S.~Ren, and J.~Sun, ``Deep residual learning for image
  recognition,'' in \emph{IEEE Conf Comput. Vision Pattern Recognit. (CVPR)},
  Jun. 2016.

\bibitem{vae}
D.~P. Kingma and M.~Welling, ``Auto-encoding variational bayes,'' in \emph{Int.
  Conf. Learn. Representations (ICLR)}, Apr. 2014.

\bibitem{gan}
I.~Goodfellow, J.~Pouget-Abadie, M.~Mirza, B.~Xu, D.~Warde-Farley, S.~Ozair,
  A.~Courville, and Y.~Bengio, ``Generative adversarial nets,'' in \emph{Adv.
  Neural Inf. Process. Syst. (NIPS)}, Dec. 2014.

\bibitem{radford}
A.~Radford, L.~Metz, and S.~Chintala, ``Unsupervised representation learning
  with deep convolutional generative adversarial networks,'' in \emph{Int.
  Conf. Learn. Representations (ICLR)}, May 2016.

\bibitem{unfold}
K.~Gregor and Y.~LeCun, ``Learning fast approximations of sparse coding,'' in
  \emph{Int. Conf Mach. Learn. (ICML)}, Jun. 2010, pp. 399--406.

\bibitem{learnwmmse}
H.~Sun, X.~Chen, Q.~Shi, M.~Hong, X.~Fu, and N.~D. Sidiropoulos, ``Learning to
  optimize: Training deep neural networks for interference management,''
  \emph{IEEE Trans. Signal Process.}, vol.~66, no.~20, pp. 5438--5453, Aug.
  2018.

\bibitem{jsac}
W.~Cui, K.~Shen, and W.~Yu, ``Spatial deep learning for wireless scheduling,''
  \emph{IEEE J. Sel. Areas Commun.}, vol.~37, pp. 1248--1261, Jun. 2019.

\bibitem{szegedy}
C.~Szegedy, W.~Zaremba, I.~Sutskever, J.~Bruna, D.~Erhan, I.~Goodfellow, and
  R.~Fergus, ``Intriguing properties of neural networks,'' in \emph{Int. Conf.
  Learn. Representations (ICLR)}, Apr. 2014.

\bibitem{fawzi}
A.~Fawzi, S.~Moosavi-Dezfooli, and P.~Frossard, ``Robustness of classifiers:
  from adversarial to random noise,'' in \emph{Adv. Neural Inf. Process. Syst.
  (NIPS)}, Dec. 2016, pp. 1632--1640.

\bibitem{natarajan}
N.~Natarajan, I.~S. Dhillon, P.~Ravikumar, and A.~Tewari, ``Learning with noisy
  labels,'' in \emph{Adv. Neural Inf. Process. Syst. (NIPS)}, Dec. 2013.

\bibitem{joulin}
A.~Joulin, L.~van~der Maaten, A.~Jabri, and N.~Vasilache, ``Learning visual
  features from large weakly supervised data,'' in \emph{Eur. Conf. Comput.
  Vision (ECCV)}, Nov. 2015.

\bibitem{veit}
A.~Veit, N.~Alldrin, G.~Chechik, I.~Krasin, A.~Gupta, and S.~Belongie,
  ``Learning from noisy large-scale datasets with minimal supervision,'' in
  \emph{IEEE Conf Comput. Vision Pattern Recognit. (CVPR)}, Jul. 2017, pp.
  839--847.

\bibitem{rolnick}
D.~Rolnick, A.~Veit, S.~Belongie, and N.~Shavit, ``Deep learning is robust to
  massive label noise,'' 2017, [Online] Available:
  https://arxiv.org/pdf/1705.10694.pdf.

\bibitem{sinha}
A.~Sinha, H.~Namkoong, R.~Volpi, and J.~Duchi, ``Certifying some distributional
  robustness with principled adversarial training,'' 2017, [Online] Available:
  https://arxiv.org/pdf/1710.10571.pdf.

\bibitem{pleiss}
G.~Pleiss, M.~Raghavan, F.~Wu, J.~Kleinberg, and K.~Q. Weinberger, ``On
  fairness and calibration,'' in \emph{Adv. Neural Inf. Process. Syst. (NIPS)},
  Dec. 2017.

\bibitem{duchi}
J.~Duchi and H.~Namkoong, ``Variance-based regularization with convex
  objectives,'' \emph{J. Mach. Learn. Res. (JMLR)}, vol.~20, no.~1, pp.
  2450--2504, Jan. 2019.

\bibitem{hein}
M.~Hein, M.~Andriushchenko, and J.~Bitterwolf, ``Why relu networks yield
  high-confidence predictions far away from the training data and how to
  mitigate the problem,'' in \emph{IEEE Conf. Comput. Vision Pattern Recognit.
  (CVPR)}, Jun. 2019, pp. 41--50.

\bibitem{jianping}
J.~Fan, Y.~Shen, N.~Zhou, and Y.~Gao, ``Harvesting large-scale weakly-tagged
  image databases from the web,'' in \emph{IEEE Conf. Comput. Vision Pattern
  Recognit. (CVPR)}, Jun. 2010.

\bibitem{ordonez}
V.~Ordonez, G.~Kulkarni, and T.~Berg, ``Im2text: Describing images using 1
  million captioned photographs,'' in \emph{Adv. Neural Inf. Process. Syst.
  (NIPS)}, Dec. 2011, pp. 1143--1151.

\bibitem{yanxia}
Y.~Xia, X.~Cao, F.~Wen, and J.~Sun, ``Well begun is half done: Generating
  high-quality seeds for automatic image dataset construction from web,'' in
  \emph{Eur. Conf. Comput. Vision (ECCV)}, Sep. 2014, pp. 387--400.

\bibitem{shixiang}
S.~Gu and L.~Rigazio, ``Towards deep neural network architectures robust to
  adversarial examples,'' in \emph{Int. Conf. Learn. Representations (ICLR)},
  May 2015.

\bibitem{bishop}
C.~M. Bishop, ``Training with noise is equivalent to tikhonov regularization,''
  \emph{Neural Comput.}, vol.~7, no.~1, pp. 108--116, Jan. 1995.

\bibitem{neuralsmithing}
R.~D. Reed and {R. J. Marks, II}, \emph{Neural Smithing: Supervised Learning in
  Feedforward Artificial Neural Networks}.\hskip 1em plus 0.5em minus
  0.4em\relax Bradford Books, Feb. 1999.

\bibitem{shaham}
U.~Shaham, Y.~Yamada, and S.~Negahban, ``Understanding adversarial training:
  Increasing local stability of neural nets through robust optimization,''
  2015, [Online] Available: https://arxiv.org/pdf/1511.05432.pdf.

\bibitem{moosavi}
S.~Moosavi-Dezfooli, A.~Fawzi, and P.~Frossard, ``Deepfool: a simple and
  accurate method to fool deep neural networks,'' in \emph{IEEE Conf Comput.
  Vision Pattern Recognit. (CVPR)}, Jun. 2016.

\bibitem{madry}
A.~Madry, A.~Makelov, L.~Schmidt, D.~Tsipras, and A.~Vladu, ``Towards deep
  learning models resistant to adversarial attacks,'' in \emph{Int. Conf.
  Learn. Representations (ICLR)}, May 2018.

\bibitem{graves}
A.~Graves, A.~Mohamed, and G.~Hinton, ``Speech recognition with deep recurrent
  neural networks,'' in \emph{IEEE Int. Conf. Acoust., Speech, and Signal
  Process. (ICASSP)}, May 2013.

\bibitem{benpoole}
B.~Poole, J.~Sohl-Dickstein, and S.~Ganguli, ``Analyzing noise in autoencoders
  and deep networks,'' Jun. 2014, [Online] Available:
  https://arxiv.org/pdf/1406.1831.pdf.

\bibitem{caglar}
C.~Gulcehre, M.~Moczulski, M.~Denil, and Y.~Bengio, ``Noisy activation
  functions,'' in \emph{Int. Conf. Mach. Learn. (ICML)}, Jun. 2016.

\bibitem{neelakantan}
A.~Neelakantan, L.~Vilnis, Q.~V. Le, I.~Sutskever, L.~Kaiser, K.~Kurach, and
  J.~Martens, ``Adding gradient noise improves learning for very deep
  networks,'' Nov. 2015, [Online] Available:
  https://arxiv.org/pdf/1511.06807.pdf.

\bibitem{mozhou}
M.~Zhou, T.~Liu, Y.~Li, D.~Lin, E.~Zhou, and T.~Zhao, ``Towards understanding
  the importance of noise in training neural networks,'' in \emph{Int. Conf.
  Mach. Learn. (ICML)}, Jun. 2019.

\bibitem{FlashLinQ}
X.~Wu, S.~Tavildar, S.~Shakkottai, T.~Richardson, J.~Li, R.~Laroia, and
  A.~Jovicic, ``{FlashL}in{Q}: {A} synchronous distributed scheduler for
  peer-to-peer ad hoc networks,'' \emph{{\it IEEE/ACM Trans. Netw.}}, vol.~21,
  no.~4, pp. 1215--1228, Aug. 2013.

\bibitem{shen_ISIT17}
K.~Shen and W.~Yu, ``{FPLinQ: A} cooperative spectrum sharing strategy for
  device-to-device communications,'' in \emph{IEEE Int. Symp. Inf. Theory
  (ISIT)}, Jun. 2017, pp. 2323--2327.

\bibitem{luo_TSP11}
Q.~Shi, M.~Razaviyayn, Z.-Q. Luo, and C.~He, ``An iteratively weighted {MMSE}
  approach to distributed sum-utility maximization for a {MIMO} interfering
  broadcast channel,'' \emph{{\it IEEE Trans. Signal Process.}}, vol.~59,
  no.~9, pp. 4331--4340, Apr. 2011.

\bibitem{ITLinQ}
N.~Naderializadeh and A.~S. Avestimehr, ``{ITL}in{Q}: {A} new approach for
  spectrum sharing in device-to-device communication systems,'' \emph{{\it IEEE
  J. Sel. Areas Commun.}}, vol.~32, no.~6, pp. 1139--1151, Jun. 2014.

\bibitem{Guo_TCOM17}
B.~Zhuang, D.~Guo, E.~Wei, and M.~L. Honig, ``Scalable spectrum allocation and
  user association in networks with many small cells,'' \emph{{\it IEEE Trans.
  Commun.}}, vol.~65, no.~7, pp. 2931--2942, Jul. 2017.

\bibitem{color}
I.~Rhee, A.~Warrier, J.~Min, and L.~Xu, ``{DRAN: D}istributed randomized {TDMA}
  scheduling for wireless ad hoc networks,'' \emph{{\it IEEE Trans. Mobile
  Comput.}}, vol.~8, no.~10, pp. 1384--1396, Oct. 2009.

\bibitem{MAPEL}
L.~P. Qian and Y.~J. Zhang, ``{S-MAPEL: M}onotonic optimization for non-convex
  joint power control and scheduling problems,'' \emph{{\it IEEE Trans.
  Wireless Commun.}}, vol.~9, no.~5, pp. 1708--1719, May 2010.

\bibitem{Johansson_TWC06}
M.~Johansson and L.~Xiao, ``Cross-layer optimization of wireless networks using
  nonlinear column generation,'' \emph{{\it IEEE Trans. Wireless Commun.}},
  vol.~5, no.~2, pp. 435--445, Feb. 2006.

\bibitem{shenouda}
M.~B. Shenouda and T.~N. Davidson, ``Convex conic formulations of robust
  downlink precoder designs with quality of service constraints,'' \emph{IEEE
  J. Sel. Topics Signal Process.}, vol.~1, no.~4, pp. 714--724, Dec. 2007.

\bibitem{monowar}
M.~Hasan, E.~Hossain, and D.~I. Kim, ``Resource allocation under channel
  uncertainties for relay-aided device-to-device communication underlaying
  {LTE-A} cellular networks,'' \emph{{\it IEEE Trans. Wireless Commun.}},
  vol.~13, no.~4, pp. 2322--2338, Mar. 2014.

\bibitem{shenkwak}
Y.~Shen and K.~S. Kwak, ``Robust power control for cognitive radio networks
  with proportional rate fairness,'' \emph{ICT Express}, vol.~1, pp. 22--25,
  June 2015.

\bibitem{weihua}
W.~Wu, R.~Liu, Q.~Yang, and T.~Q.~S. Quek, ``Learning-based robust resource
  allocation for {D2D} underlaying cellular network,'' \emph{IEEE Trans.
  Wireless Commun.}, vol.~21, no.~8, pp. 6731--6745, Aug. 2022.

\bibitem{junwang}
J.~Wang, J.~Chen, Y.~Lu, M.~Gerla, and D.~Cabric, ``Robust power control under
  location and channel uncertainty in cognitive radio networks,'' \emph{{\it
  IEEE Wireless Commun. Lett.}}, vol.~2, no.~4, pp. 113--116, April 2015.

\bibitem{dallanese}
E.~Dall’Anese, S.~Kim, G.~B. Giannakis, and S.~Pupolin, ``Power control for
  cognitive radio networks under channel uncertainty,'' \emph{{\it IEEE Trans.
  Wireless Commun. }}, vol.~10, no.~10, pp. 3541 -- 3551, August 2011.

\bibitem{chalise}
B.~K. Chalise, S.~Shahbazpanahi, A.~Czylwik, and A.~B. Gershman, ``Robust
  downlink beamforming based on outage probability specifications,'' \emph{IEEE
  Trans. Wireless Commun.}, vol.~6, no.~10, pp. 3498--3503, Oct. 2007.

\bibitem{shenouda2}
M.~B. Shenouda and T.~N. Davidson, ``Probabilistically-constrained approaches
  to the design of the multiple antenna downlink,'' in \emph{Asilomar Conf.
  Signals Syst. Comput.}, Oct. 2008.

\bibitem{kunyu}
K.~Wang, A.~M. So, T.~Chang, W.~Ma, and C.~Chi, ``Outage constrained robust
  transmit optimization for multiuser {MISO} downlinks: Tractable
  approximations by conic optimization,'' \emph{IEEE Trans. Signal Process.},
  vol.~62, no.~21, pp. 5690--5705, Nov. 2014.

\bibitem{foad}
F.~Sohrabi and T.~N. Davidson, ``Coordinate update algorithms for robust power
  loading for the {MU-MISO} downlink with outage constraints,'' \emph{IEEE
  Trans. Signal Process.}, vol.~64, no.~11, pp. 2761--2773, Jan. 2016.

\bibitem{medra}
M.~Medra and T.~N. Davidson, ``Per-user outage-constrained power loading
  technique for robust {MISO} downlink,'' in \emph{Asilomar Conf. Signals Syst.
  Comput.}, Nov. 2015.

\bibitem{elnourani}
M.~Elnourani, S.~Deshmukh, B.~Beferull-Lozano, and D.~Romero, ``Robust underlay
  device-to-device communications on multiple channels,'' Feb. 2020, [Online]
  Available: https://arxiv.org/pdf/2002.11500.pdf.

\bibitem{kerret}
P.~de~Kerret and D.~Gesbert, ``Robust decentralized joint precoding using team
  deep neural network,'' in \emph{Int. Symp. Wireless Commun. Syst. (ISWCS)},
  Aug. 2018.

\bibitem{junbeom}
J.~Kim, H.~Lee, and S.~Park, ``Learning robust beamforming for {MISO} downlink
  systems,'' \emph{IEEE Commun. Lett.}, vol.~25, no.~6, pp. 1916--1920, Mar.
  2021.

\bibitem{runze}
R.~Dong, B.~Wang, and K.~Cao, ``Deep learning driven {3D} robust beamforming
  for secure communication of {UAV} systems,'' \emph{IEEE Wireless Commun.
  Lett.}, vol.~10, no.~8, pp. 1643--1647, Aug. 2021.

\bibitem{rayleigh}
B.~Sklar, ``Rayleigh fading channels in mobile digital communication systems
  {Part I: Characterization},'' \emph{IEEE Commun. Mag.}, vol.~35, no.~7, pp.
  90--100, Jul. 1997.

\bibitem{rci}
C.~B. Peel, B.~M. Hochwald, and A.~L. Swindlehurst, ``A vector-perturbation
  technique for near-capacity multiantenna multiuser communication-part i:
  channel inversion and regularization,'' \emph{IEEE Trans. Commun.}, vol.~53,
  no.~1, pp. 195--202, Jan. 2005.

\bibitem{sgd}
J.~Kiefer and J.~Wolfowitz, ``Stochastic estimation of the maximum of a
  regression function,'' \emph{Ann. Math. Statist.}, vol.~23, no.~3, pp.
  462--466, Sep. 1952.

\bibitem{evolution}
D.~E. Goldberg and J.~H. Holland, ``Genetic algorithms and machine learning,''
  \emph{Mach. Learn.}, vol.~3, pp. 95--99, Oct. 1988.

\bibitem{chenyu}
C.~Lee, S.~Xie, P.~Gallagher, Z.~Zhang, and Z.~Tu, ``Deeply-supervised nets,''
  in \emph{Int. Conf. Artif. Intell. Statist. (AISTATS)}, May 2015.

\bibitem{pytorch}
A.~Paszke \emph{et~al.}, ``{PyTorch}: An imperative style, high-performance
  deep learning library,'' in \emph{Adv. Neural Inf. Process. Syst. (NeurIPS)},
  Dec. 2019, pp. 8024--8035.

\bibitem{tensorflow}
\BIBentryALTinterwordspacing
M.~Abadi \emph{et~al.}, ``{TensorFlow}: Large-scale machine learning on
  heterogeneous systems,'' 2015, software available from tensorflow.org.
  [Online]. Available: \url{https://www.tensorflow.org/}
\BIBentrySTDinterwordspacing

\bibitem{vanzwet}
W.~R.~V. Zwet, \emph{Convex Transformations of Random Variables}.\hskip 1em
  plus 0.5em minus 0.4em\relax Amsterdam, Netherlands: Mathematisch Centrum,
  1964, ch.~3.

\bibitem{chatillon}
G.~Chatillon, R.~Gelinas, L.~Martin, and L.~Laurencelle, ``When is it
  preferable to estimate population percentiles from a set of classes rather
  than from the raw data?'' \emph{J. Educ. Statist.}, vol.~12, no.~4, pp.
  395--409, 1987.

\bibitem{mmse_channel}
H.~V. Poor, \emph{An Introduction to Signal Detection and Estimation},
  2nd~ed.\hskip 1em plus 0.5em minus 0.4em\relax New York, NY, USA: Springer
  Science \& Business Media, Feb. 1994.

\bibitem{mung}
M.~Chiang, C.~W. Tan, D.~P. Palomar, D.~O'Neill, and D.~Julian, ``Power control
  by geometric programming,'' \emph{IEEE Trans. Wireless Commun.}, vol.~6,
  no.~7, pp. 2640--2651, Jul. 2007.

\end{thebibliography}

\end{document}